\documentclass{new_tlp}

\usepackage[utf8]{inputenc}

\usepackage[english]{babel}

\usepackage{amsmath,amssymb}
\usepackage{booktabs}
\usepackage{morefloats}
\usepackage{placeins}

\usepackage{microtype}

\title[Theory and Practice of Logic Programming]{Exchanging Conflict Resolution in an\\Adaptable Implementation of ACT-R}

\author[D. Gall and T. Fr\"uhwirth]{DANIEL GALL and THOM FR\"UHWIRTH\\
      Faculty of Engineering and Computer Science, Ulm University, Germany\\
      \email{\{daniel.gall,thom.fruehwirth\}@uni-ulm.de}}

\newtheorem{definition}{Definition} 

\newtheorem{example}{Example} 

\newcommand*{\rel}[1]{\mathord{\mathrm{#1}}}

\usepackage{tikz}
\usetikzlibrary{positioning,shapes,shapes.multipart,shadows,arrows,automata,calc,decorations.markings}      

\usepackage{url}

\begin{document}

\maketitle

\begin{abstract}
In computational cognitive science, the cognitive architecture ACT-R is 
very popular. It describes a model of cognition that is amenable to 
computer implementation, paving the way for computational psychology. Its 
underlying psychological theory has been investigated in many 
psychological experiments, but ACT-R lacks a formal definition of its 
underlying concepts from a mathematical-computational point of view. Although the 
canonical implementation of ACT-R is now modularized, this production 
rule system is still hard to adapt and extend in central components like the 
conflict resolution mechanism (which decides which of the applicable 
rules to apply next).\looseness=-1 

In this work, we present a concise implementation of ACT-R
based on Constraint Handling Rules which has been derived from a 
formalization in prior work. To show the adaptability 
of our approach, we implement several different conflict resolution 
mechanisms discussed in the ACT-R literature. This results in the first 
implementation of one such mechanism. For the other mechanisms, we 
empirically evaluate if our implementation matches the results of 
reference implementations of ACT-R.
\end{abstract}


\section{Introduction}

Computational cognitive modeling is an approach in cognitive sciences which explores human cognition by implementing detailed computational models. This enables researchers to execute their models and simulate human behavior \cite{sun_introduction_2008}. Due to their executability, computational models have to be defined precisely. Thereby ambiguities appearing in verbal-conceptual models can be eliminated. By conducting the same experiments with humans and an executable cognitive model, the plausibility of a model can be verified and gradually improved.

To implement cognitive models, it is helpful to introduce \emph{cognitive architectures} which bundle well-investigated research results from several disciplines of psychology to a unified theory. On the basis of such an architecture, researchers are able to implement domain-specific computational models without having to deal with the remodeling of fundamental psychological results. Additionally, cognitive architectures ideally constrain modeling to plausible models which facilitates the modeling process \cite{taatgen_modeling_2006}.

One of the most popular cognitive architectures is \emph{Adaptive Control of Thought -- Rational} (ACT-R), a production rule system introduced by John R. Anderson \cite{AndersonLe98,anderson_integrated_2004}. It has been used to model cognitive tasks like learning the past tense \cite{taatgen_why_2002}, but is also used in human-computer interaction or to improve educational software by simulating human students  \cite[p. 1045 sqq.]{anderson_integrated_2004}. Although providing a theory of the psychological foundations, ACT-R lacks a formal definition of its underlying concepts from a mathematical-computational point of view. This led to a reference implementation full of assumptions and technical artifacts beyond the theory making it difficult to overlook and inhibiting adaptability and extensibility. The situation improved with the modularization of the psychological theory, but it is still difficult to exchange more central parts of the implementation like conflict resolution \cite{stewart_deconstructing_2007}.

To overcome these drawbacks, we have formalized parts of the implementation closing the gap between the psychological theory and the technical implementation. We describe an implementation of ACT-R which has been derived from our formalization using Constraint Handling Rules (CHR). Due to the power of logic programming, our implementation is very close to the formalization and leads to short and concise code covering the fundamental parts of the ACT-R theory. For the compilation of ACT-R models to CHR programs, source-to-source transformation is used. Our implementation is highly adaptable. In this paper, this is demonstrated by integrating four different conflict resolution strategies. Despite its proximity to the theory, the implementation can reproduce the results of the original implementation as exemplified in the evaluation of our work. The formalization may support the understanding of the details of our implementation, hence we refer to \cite{gall_rule_based_2013} and and the online appendix (\ref{sec:formalization}).

%

In section~\ref{sec:actr}, we give an overview of the fundamental concepts of ACT-R and shortly describe their implementation in CHR. Section~\ref{sec:conflict_resolution} describes the general conflict resolution process of ACT-R. Then the implementation of four different conflict resolution strategies proposed in the literature is presented. To evaluate our implementations, we use an example to compare the results of our implementation with those of the reference implementations where available in section~\ref{sec:evaluation}. Eventually, in section~\ref{sec:related_work} some related work is presented and a conclusion is given in section~\ref{sec:conclusion}.


\section{A CHR implementation of ACT-R}
\label{sec:actr}

In the following, a short overview of the fundamental concepts of the ACT-R theory and their transfer to CHR is given. For reasons of space, we refer to the literature for an introduction to CHR \cite{fru_chr_book_2009}. For a more detailed introduction to ACT-R, see \cite{anderson_integrated_2004} and \cite{taatgen_modeling_2006}. The reference implementation of ACT-R is written in Lisp and can be obtained from the ACT-R website \cite{actr_homepage}. Details of our implementation including the formalization it is based on can be found in~\cite{gall_rule_based_2013}. Parts of the formalization are located in the online appendix (\ref{sec:formalization}).

\subsection{Architecture}
\label{sec:modular_architecture}

ACT-R is a production rule system which distinguishes two types of knowledge: \emph{declarative knowledge} holding static facts and \emph{procedural knowledge} representing processes controlling human cognition. For example, in a model of the game \emph{rock, paper, scissors}, a declarative fact could be ``The opponent played scissors'', whereas a procedural information could be that a round is won, if we played rock and the opponent played scissors. Declarative knowledge is represented as \emph{chunks}. Each chunk consists of a symbolic name and labeled slots which hold symbolic values. The values can refer to other chunk names, i.e. chunks can be connected. Chunks are typed, i.e. the number and names of the slots provided by a chunk are determined by a type. As usual for production rule systems, procedural knowledge is represented as rules of the form IF \textit{conditions} THEN \textit{actions}. Conditions match values of chunks, actions modify them.

\looseness=-1 The psychological theory of ACT-R is modular: 
There are modules for each function of the human mind like a declarative module holding the declarative facts, a goal module taking track of the current goal of a task and buffering information and a procedural module holding the procedural information and controlling the cognitive process. There are also modules to interact with the environment like a visual module perceiving the visual field. 
The modules are independent from each other, i.e. there is no direct communication between them. Each module has a fixed number of \emph{buffers} associated with it. The buffers can hold at most one single piece of information a time, i.e. one chunk. Modules can put chunks into their associated buffers.

%
%
%
%
%
%
%
%
%

The core of the system is the procedural module which can access the buffers of all other modules but does not have an own buffer. It consists of a \emph{procedural memory} with a set of production rules. The conditions of a production rule refer to the contents of the buffers, i.e. they match the values of the chunk's slots. The formal applicability condition of rules can be found in the online appendix (\ref{sec:formalization}).  

There are three types of actions whose arguments are encoded as chunks as well: 
 First of all, \emph{buffer modifications} change the content of a buffer, i.e. the values of some of the slots of a chunk in a buffer. Secondly, the procedural module can state \emph{requests} to other modules which then change the contents of their buffers.  Eventually, \emph{buffer clearings} remove the chunk from a buffer. 
 Although our implementation can handle requests and clearings, we only regard buffer modifications in this work for the sake of simplicity.

\begin{example}
\label{ex:simple_rule}
Consider the following rule:
\begin{verbatim}
(p recognize-win  
   =goal>  isa game   me rock   opponent scissors            
 ==>   
   =goal>  result win)     
\end{verbatim}
It recognizes a win situation in the game \emph{rock, paper, scissors} if the model has realized that the opponent played scissors and the agent played rock (which could be accomplished by a corresponding production rule interacting with the visual module). The situation is represented by a chunk of type \verb|game| providing the slots \verb|me|, \verb|opponent| and \verb|result|. As a result, it adds the information that the round has been won by modifying the \verb|result|-slot of the goal buffer. 
\end{example}

Furthermore, the procedural module controls the \emph{match-select-apply} cycle of the production rule system. It searches for matching rules. As soon as a matching rule has been selected to fire, it takes 50\,ms for the rule to fire based on theories of human cognition \cite[p. 54]{anderson_how_2007}. During this time, the matching process is inhibited and no other rule can be selected until the selected rule is applied. Hence, the productions are executed serially. The production system is called \emph{free}, if no rule is selected and waiting for execution. As long as the procedural module is free, it searches for matching rules.

The modules act in parallel. When a request is sent to a module by a production, the procedural module becomes free while the request is completed. Hence, new production rules can match while other modules might be busy with requests.

ACT-R can be extended by arbitrary modules communicating through buffers with the procedural system. However, to exchange more fundamental parts of the architecture it needs more than only architectural modules as shown in section~\ref{sec:conflict_resolution}.

\subsection{The Procedural Module in CHR}

The procedural module is the core of ACT-R's production rule system. Our implementation is based on the translation of production rule systems to CHR as presented in \cite[chapter 6.1]{fru_chr_book_2009}. However, we have to account for the concepts of chunks and buffers, since ACT-R differs in those particular points from other production systems. Details of the implementation can be found in \cite{gall_rule_based_2013}.

The set of chunks can be represented in CHR by a constraint \verb|chunk(C,T)|, where \verb|C| is the name of the chunk and \verb|T| its type. The slots provided by this chunk and their values can be stored in constraints \verb|chunk_has_slot(C,S,V)| denoting that chunk \verb|C| has the value \verb|V| in slot \verb|S|. With special consistency rules it can be assured, that no chunk has two values in its slots and that it only provides the slots allowed by its type. Analogously, a buffer is represented by a constraint \verb|buffer(B,M,C)| denoting that the buffer \verb|B| is affiliated with the module \verb|M| and holds chunk \verb|C|. The formal definitions of chunks and buffers can be found in the online appendix (\ref{sec:formalization}).

A production rule can now match and modify the information of the buffer system. The actions are implemented by trigger constraints \verb|buffer_action(B,C)| which get the name of the buffer \verb|B| and a chunk description \verb|C| represented by a term \verb|chunk(C,T,[(S,V),...])| which describes a chunk with name \verb|C|, type \verb|T| and a list of slot-value pairs representing the values of the chunk's slots. Note that such chunk descriptions can be incomplete in some arguments by simply letting them unspecified.

\begin{example}
\label{ex:simple_rule_chr}
The rule from example~\ref{ex:simple_rule} can be translated to the following CHR rule:
\begin{verbatim}
buffer(goal,_,C), chunk(C,game), 
  chunk_has_slot(C,me,rock), chunk_has_slot(C,opponent,scissors) 
     ==> buffer_modification(goal,chunk(_,_,[(result,win)])).                                                                            
\end{verbatim}
The name and type of the chunk in the modification are not specified in the original rule and therefore left blank as well as the \verb|me| and \verb|opponent| slots.
\end{example}

\subsection{Timing and Phases}
\label{sec:timing}

As mentioned before, the production system of ACT-R is occupied for 50\,ms after a rule has been selected. To model such latencies, an event queue has to be added. It keeps track of the current time and holds an ordered set of events which can be dequeued one after another according to their scheduled times. In our implementation, the event queue is implemented as a priority queue sorting its elements after the time and a priority determining the order of application for simultaneous events. Events are arbitrary Prolog goals and can be added by \verb|add_q(Time,Priority,Event)|. The current time can be queried by \verb|get_time(Now)|.

To ensure that a production rule only matches when the module is free, we replace each CHR rule of the form \verb!C ==> A! according to the following scheme consisting of two rules:
\begin{verbatim}
C \ match <=> add_q(Now + 0.05,0,apply_rule(rule(r,C))).
C \ apply_rule(rule(r,C)) <=> A, get_time(Now), add_q(Now,-10,match).
\end{verbatim}
The constraint \verb|match| indicates that the procedural module is free and searches for a matching rule. For the matching rule, an \verb|apply_rule| event is scheduled 50\,ms from the current time. This event will actually fire the rule. The actions \verb|A| schedule their effects on the buffers at the current time with different priorities. Requests are only sent to the corresponding module. Its effects on the requested buffer are scheduled at a later time. Finally, a new \verb|match| event is scheduled at the current time \verb|Now| but with low priority of $-10$. This ensures that all current actions are performed before the next rule is scheduled to fire.\looseness=-1

Otherwise, if no rule matches and the procedural module is free (i.e. a \verb|match| constraint is present), a rule can only become matching if the content of the buffers change. Hence, a new \verb|match| constraint is added directly after the next event in the queue. This models the fact that the procedural module is searching permanently for matching rules when it is free without adding unnecessary \verb|match| events.

\section{Conflict Resolution}
\label{sec:conflict_resolution}

Only one matching production rule can fire at a time. Hence, if there are multiple applicable productions, the system has to decide which to fire. This process is called \emph{conflict resolution} \cite{mcdermott_1977}. In most implementations, CHR simply chooses the rule to fire by textual order, which is a valid conflict resolution mechanism. However, in ACT-R a more advanced approach using subsymbolic concepts is needed to faithfully model human cognition.

\subsection{General Conflict Resolution Process}

In \cite[p. 151]{fru_chr_book_2009} a general method to implement different conflict resolution mechanisms in CHR is given. This method is adapted to our CHR implementation of ACT-R. The first rule of each CHR rule pair from section~\ref{sec:timing} can be replaced by:
\begin{verbatim}
match, C ==> G | conflict_set(rule(r,C)).
\end{verbatim}
Hence, the application of a matching production is delayed by adding the rule to the conflict set instead of choosing the first matching rule to be applied by scheduling \verb|apply_rule/1| as explained in section~\ref{sec:timing}. Thereby all matching rules are collected in \verb|conflict_set/1| constraints which then can be reduced to one single constraint containing only the rule to be applied according to an arbitrary strategy. 

As a last production rule, the rule \verb!match <=> select.! occurs in the program. This rule will always be applied last (since rules are applied in textual order in CHR). It removes the remaining \verb|match| constraint and adds a constraint \verb!select! which triggers the selection process. This means that the conflict resolution is performed by choosing one rule from the conflict set constraints and removing all other such constraints. If no rule matches, a new \verb|match| constraint is scheduled after the next event. 

With the introduction of the \verb|select| constraint, the system commits to the rule to be applied by scheduling the corresponding \verb!apply_rule/1! event as explained in section~\ref{sec:timing}. This leads the chosen production to perform its actions since its second CHR rule is applicable. After the actions are performed, the next matching phase is scheduled.

The strategy of how the conflict set is eliminated to one single rule which will be applied may vary and is exchangeable. In the following section, several strategies are presented and implemented.

\subsection{Conflict Resolution Strategies}

There have been several conflict resolution strategies proposed for ACT-R over time. To demonstrate the adaptability of our CHR implementation, we implement some of those strategies. In the reference implementation of ACT-R, such adaptations might need a lot of knowledge about its internal structures \cite{stewart_deconstructing_2007}.

In general, ACT-R conflict resolution strategies usually use the subsymbolic concept of \emph{production utilities}. The production utility for a production $i$ is the function $U_i: \mathbb{N} \rightarrow \mathbb{R}$ which expresses the value of utility of a particular production at its $n$th application which may be adapted according to a learning strategy. In the conflict resolution process, the current utility values  are compared for all matching functions and the production with the highest utility is chosen. The production utility can therefore be seen as a dynamic rule priority which is adapted according to a certain strategy.

In the following, we present some different learning strategies to adapt the utility of a production. Eventually, the concept of rule refraction is introduced, which is a general conflict resolution concept and can be applied for all of the presented learning strategies.

\subsubsection{Reinforcement-Learning-Based Utility Learning}

The current implementation of ACT-R~6.0 uses a conflict resolution mechanism which is motivated by the Rescorla-Wagner learning equation \cite{rescorla_wagner_1972}. The basic concept is that there are special production rules which recognize a successful state (by some model-specific definition) and then trigger a certain amount of reward measured in units of time as a representation of the effort a person is willing to spend to receive a certain reward \cite[p. 161]{anderson_how_2007}. All productions which lead to the successful state, i.e. all productions which have been applied, receive a part of the triggered amount of reward which demounts the more time lies between the application of the production rule and the triggering of the reward. The utility $U_i$ of a production $i$ then is adapted as follows:
\begin{equation}
\label{eq:utility_learning:reinforcement}
U_i(n) = U_i(n-1) + \alpha (R_i(n) - U_i(n-1))
\end{equation}
The reward $R_i(n)$ for the $n$th application of the rule $i$ is the difference of the external reward and the time between the selection of the rule and the triggering of the reward. The utility adapts gradually to the average reward a rule receives. Its calculation can be extended by noise to enable rules with initally low utilities to fire. This then may boost their utility values.

In CHR, this strategy can be implemented as follows: For each production rule, a \verb|utility/2| constraint is stored holding its current utility value. For rules marked with a reward, a \verb|reward/2| constraint holds the amount of reward.  When a production rule is applied, this information is stored with the application time in a constraint by the rule \verb!apply_rule(rule(P,_,_)) ==> get_time(Now), applied([(P,Now)]).! With a corresponding rule, the \verb|applied/1| constraints are merged respecting the application time of the rules, since the adaptation strategy depends on the last utility value of a rule and rules might be applied more than once until they receive a reward. This leads to one \verb|applied/1| constraint containing a sorted list of rules and their application time. 

If a rule which is marked with a reward is going to be applied, the reward can be triggered by \verb!apply_rule(rule(P,_)), reward(P,R) ==> trigger_reward(R).! The triggering of the reward simply adapts the utilities according to equation~\ref{eq:utility_learning:reinforcement} for all productions which have been applied indicated by the \verb!applied/1! constraint respecting the order of application. Afterwards, this constraint is deleted because after a reward has been received, the rule is not considered in the next adaptation.


\subsubsection{Success-/Cost-Based Utility Learning}
\label{sec:success_cost_based_utility_learning}
In prior implementations of ACT-R, the utility learning is based on a success-/cost approach \cite{anderson_integrated_2004,taatgen_modeling_2006}. A detailed description can be found in \cite[unit~6]{actr5_tutorial}. Each production rule $i$ is associated to the values $P_i$ denoting the success probability of the production and $C_i$ denoting its costs. In this approach, the utility of a production rule is defined as:
\begin{equation}
U_i(n) = P_i(n) G - C_i(n)
\end{equation}
Note that the current utility does not depend on the value of the last utility, but can be calculated by the current values of the parameters instead. Hence, the order of application does not play a role. 
Usually, $C_i$ is measured in units of time to achieve a goal whereas $G$ -- the goal value -- is an architectural parameter and usually set to 20\,s. The parameters $P$ and $C$ are obtained by the following equations:
\begin{align}
P_i(n) &= \frac{\mathit{\#sucesses_i}}{\mathit{\#successes_i + \#failures_i}} & C_i(n) &= \frac{\mathit{efforts_i}}{\mathit{\#successes_i + \#failures_i}}
\end{align}
The values $\mathit{\#sucesses}$ and $\mathit{\#failures}$ count all applications of a rule which have been identified as a success or a failure respectively. Similarly to the reinforcement-based learning, some productions which identify a success or failure trigger an event which adapts the counters of successes or failures of all production rules which have been applied since the last triggering. The efforts are estimated by the difference of the time of the triggering and the selection of a rule. The values are initialized with $\mathit{\#sucesses} = 1, \mathit{\#failures} = 0$ and $\mathit{efforts} = 0.05\,s$ which is the selection time of one firing. Analogously to the reward-based strategy, utilities can be extended by noise. 

Similarly to the implementation of the reinforcement learning rule, the triggering of a success or failure can be achieved by a constraint \verb|success(P)| or \verb|failure(P)|, which encode that a production \verb|P| is marked as success or failure respectively. Combined with the \verb|apply_rule/2| constraint, a \verb|success/0| or \verb|failure/0| constraint can be propagated which trigger the utility adaptation.  The following rules show the adaptation of $\mathit{\#successes_i}$ and $\mathit{efforts_i}$ when a success is triggered and rule $i$ has been applied before:
\begin{verbatim}
success \ applied(P,T), efforts(P,E), successes(P,S)  <=> 
  get_time(Now), efforts(P,E+Now-T), successes(P,S+1).
success <=> true.
\end{verbatim}
The number of successes or failures are stored in the respective binary constraints and if a success is triggered, they are incremented for all applied production rules and efforts are adjusted. The rules for failures are analogous. The adaptation of one of those parameters triggers the rules which replace the constraints holding the old $P_i$ and $C_i$ values by new values. When a $P_i$ or $C_i$ constraint is replaced, the calculation of the new utility value is triggered. To ensure that only one utility value is in the store, a destructive update rule is used.

\subsubsection{Random Estimated Costs}

In \cite{BelavkinR04}, a conflict resolution strategy motivated by research results in decision-making is presented. The current implementation varies slightly from this description \cite{belavkin_optimist_impl} and we stick to this most recent approach for a better comparability of the results. The strategy is based on the success-/cost-based utility learning from section~\ref{sec:success_cost_based_utility_learning} and uses the same subsymbolic information (the counts of successes and failures and the efforts). However, instead of calculating the average cost $C_i$, the expected costs $\theta_i$ of achieving a success by a rule are estimated:
\begin{equation}
\theta_i := \mathrm{E}(C_i) \approx \frac{\mathit{efforts_i}}{\mathit{\#sucesses_i}}
\end{equation}
From the expected costs $\theta_i$ of a rule $i$, the \emph{random estimated costs} $\zeta_i$ are derived by 
by drawing a random number $r_i$ from a uniform distribution $U(0,1)$ and setting $\zeta_i = -\theta_i \cdot \mathrm{log}(1 - r_i)$. Eventually, production utilities are calculated analogously to the success-/cost-based strategy: $U_i = P_i G - \zeta_i$. The influence of the random estimated costs can be varied by adapting the parameter $G$. If $G = 0$, the production rule with minimal random estimated costs will be fired (as suggested in \cite{BelavkinR04}).

Since this method uses the same parameters as the success-/cost-based variant, almost all of the code can be reused for an implementation. However, instead of the costs, the expected costs $\theta_i$ are computed and saved in a constraint whenever the success/failure ratio changes. Additionally, the random costs must be calculated in every conflict resolution step and not only when the parameters change since they vary each time due to randomization. Hence, a rule must be added which calculates the utility value as soon as a production rule enters the conflict set:
\begin{verbatim}
conflict_set(rule(P,_)), theta(P,T), succ_prob(P,SP) ==>  
  random(R), Z is -T * log(1 - R), U is SP*20-Z, 
  set_utility(P,U).
\end{verbatim}
The rest of the implementation like the calculation of the success/failure counters, efforts or the pruning of the conflict set is identical to the success-/cost-based strategy.

\subsubsection{Production Rule Refraction}

In contrast to the previous strategies which only exchange the utility learning part, production rule refraction adapts the general conflict resolution mechanism and can be combined with all of the other presented strategies. It was first suggested in \cite{young_refraction_2003} to avoid over-programming of models in the sense that the order of application of a set of rules is fixed in advance by adding artificial signals to ensure the desired order. Rule refraction can avoid such operational concepts by inhibiting the application of the same rule instantiation more than once. 
To the best of our knowledge, our implementation is the first of its kind for ACT-R.



Refraction can be implemented by saving the instantiation of each applied production using the rule \verb!apply_rule(R) ==> instantiation(R).! When building the conflict set, the following rule eliminates all productions which already have been applied from the set: \verb!instantiation(R) \ conflict_set(R) <=> true.! This pruning rule must be performed before the rule selection process, so that such productions are never considered as fire candidates.


%
%
%

\section{Evaluation}
\label{sec:evaluation}

After having implemented some different conflict resolution strategies, we test their validity with an example model of the game \emph{rock, paper, scissors}. The idea is that the model simulates a player playing against three opponents with different preferences on the three choices in the game. We then want to observe, how the model adapts its strategy under the different conflict resolution mechanisms and test if the results of the ACT-R implementation and our CHR implementation match.

\subsection{Setup}

The player is basically modeled by the production rules \verb|play-rock|, \verb|play-paper| and \verb|play-scissors| standing for the three choices a player has in the game. At the beginning, the production rules have equal utilities which are then adapted by the utility learning mechanisms of the three conflict resolution strategies. Since we only want to test our conflict resolution implementations, we try to rule out all other factors which could influence the behavior of our model. Hence, we only use the procedural module with the goal buffer and do not simulate any declarative knowledge or even perceptual and motor modules. I.e. the model is not a realistic psychological hypothesis of the game play, but only a test of our implementation. Furthermore, we disable noise where possible to better compare our results. In ACT-R, the canonical parameter setting is not recommended to change without justification \cite[sec. 1.1]{stewart_deconstructing_2007}. For our experiment, we used this setting.

The moves of the opponents are randomly generated in advance according to their defined preferences: Player~1 simply chooses rock for every move, player 2 chooses only between rock and paper and player 3 chooses equally between all three possibilities. For each player, we produced 20~samples of 20~moves (except for player~1 with only one sample of 20~moves). Their choices are put into the goal buffer one after another by host-language instructions (Lisp and Prolog/CHR). The game is played for 20~rounds until a restart with a new sample which corresponds to 2\,s simulation time. Finally, the utility values $U_{\{r,p,s\}}$ at the end of each run (for rock, paper and scissors respectively) are collected and compared to the reference implementation. We use the notation $\overline{U}_{\{r,p,s\}}$ to denote the average of those values over all 20 samples. In the following the implementation of the production rule \verb|play-rock|:
\begin{verbatim}
(p play-rock
   =goal>  isa game   me nil     opponent nil 
 ==>  
   =goal>             me rock    opponent =x   !output! (rock =x) )
\end{verbatim}
This rule simply puts the symbol \verb|rock| into the goal buffer indicating that the model chose rock. The variable \verb|=x| is set by built-in functions of the host language (omitted in the listing) modeling the choice of the opponent derived from a given list of moves. The rules for \emph{paper} and \emph{scissors} can be defined analogously. The model has been translated to CHR by our compiler. We performed the translation of Lisp built-ins to Prolog built-ins by hand. \looseness=-1

Furthermore, the model contains production rules detecting a win, draw or defeat situation (similar to example~\ref{ex:simple_rule}) and resetting the choices of the two players in the goal buffer to indicate that the next round begins. Those rules are marked with a reward (positive or negative) or as a success/failure respectively. In the case of a draw, no reward, success or failure will be triggered. Hence, the utility learning algorithms will adapt the values of the fired rules depending on their success. 

If the highest utilities in the conflict set are equal, the strategy of ACT-R is undocumented. It depends on the order of the rules in the source code and may vary between the implementations (e.g. the strategy of ACT-R~6.0 differs from ACT-R 5.0 as we found in our experiments). We adapted the order of rules in our translated CHR model to match the strategy of ACT-R. Usually, noise would rule out such differences. 

For the reference implementations, we used Clozure Common Lisp version 1.9-r15757. The CHR implementation has been run on SWI-Prolog version 6.2.6. The relevant data collected in our experiments can be found in the online appendix (\ref{sec:evaluation_results}).

\subsection{Availability of the Strategies}

Our approach enables the user to exchange the complete conflict resolution strategy without relying on provided interfaces and hooks except for the very basic information that a rule is part of the conflict set or about to be applied. This information relies on the fundamental concept of the match-select-apply cycle of ACT-R. In the reference implementations of the strategies, there are deeper dependencies and assumptions on when and how subsymbolic information is adapted and stored.

This leads to incompatibilities: The reinforcement-learning-based strategy is only available for ACT-R~6.0. Although the success-/cost-based strategy is shipped with ACT-R~6.0, it was not executable for us and hence we had to use ACT-R 5.0 to run it. This leads to further incompatibility problems when using modules not available for ACT-R 5.0 (which is in general difficult to extend due to the lack of architectural modules). Since the method of random-estimated costs relies on the success-/cost-based strategy, it is also only available for ACT-R 5.0.

Our implementation of the refraction-based method is to the best of our knowledge the only existing implementation for ACT-R, although it has been suggested in \cite{young_refraction_2003}.

\subsection{Reinforcement-Learning-Based Utility Learning}

For the reinforcement-learning-based strategy, we marked the win-detecting production rules with a reward of 2 and the defeat-detecting rules with 0 which leads to negative rewards for all applied rules when a defeat is detected. Draws do not lead to adjustments of the strategy in our configuration. We executed the model on ACT-R~6.0 version~1.5-r1451 and our CHR implementation.

Our implementation matches the results of the reference implementation exactly when rounded to the same decimal precision (see online appendix~\ref{sec:evaluation_results_reinf}). Differences of floating point precision did not influence the results, since ACT-R does round the final results to the one-thousandths. As expected, the model usually rewards the paper rule most  when playing against player~1~and~2 (average utility at end of round for player 1: $(\overline{U_r}, \overline{U_p}, \overline{U_s}) = (0, 1.87, -0.02)$; player 2: (0, 0.81, 0.49)). Exceptions are rounds where the opponent chooses paper above average especially as first moves (e.g. sample 10: 75\% rate of paper; first 9 moves; $U_p = 0, U_s = 1.329$). In such cases, scissors has the highest utility. This is reinforced by the relatively high reward of successes compared to the punishment of defeats. However, the winning rate is still very high (15 wins, 5 defeats, no draws). Overall, the behavior of the model is very successful (average: 10.4 wins, 3.9 draws and 5.7 defeats in each sample). For player~3 -- as expected -- no unique result can be learned; wins, draws and defeats are very close in average (6.6 wins, 6.7 draws, 6.7 defeats).

\subsection{Success-/Cost-Based Utility Learning}
\label{sec:evaluation:success_cost_based}

For the success-/cost-based strategy, the production rules recognizing a win situation are marked as a success and analogously the production rules for the defeat situations as a failure. We used ACT-R 5.0 to test our implementation against the reference implementation, since it is not available for ACT-R~6.0. Again, noise is disabled for better comparability. Because the selection mechanism for rules with same utility differs from ACT-R~6.0, we adapted the order in which the rules appear in the source code. 

Our implementation matches the results of the reference implementation exactly (see online appendix~\ref{sec:evaluation_results_succ}). It can be seen that this strategy is not able to detect the optimal moves for player~1. Analyses showed that due to the order of the rules, the model first selects to play \emph{rock}. This leads to a draw and hence no adaptation of the utilities. Hence, rock is played repeatedly. In real-world models, noise would help to overcome such problems. For player 2, the model correctly chose to play \emph{paper} in average even for the samples where the opponent chooses paper more often than rock. However, in average, the model did only win 8.9 out of 20 rounds in a sample and produced 9.1 draws. For each of the samples, only two rounds were lost. 

\subsection{Random Estimated Costs}

Due to the randomness of this strategy, no exact matches of results can be expected. Hence, we executed the models on 3 samples (the first of each opponent) with 50~runs for each sample. The reference implementation has been run on ACT-R~5.0. 

The average utilities are close to the reference implementation (error squares of average utilities player~1: ($\Delta\overline{U_r}^2,\Delta\overline{U_p}^2,\Delta\overline{U_s}^2) = (0.145, 0.000, 0.000)$; player~2: (0.850, 0.000, 0.098); player~3: (2.823, 0.503, 0.003), see online appendix~\ref{sec:evaluation_results_rand} for details). It can be seen that for most runs the production with the highest, medium and lowest utility value coincide. For player 1, the random estimated costs overcome the problem of the success-/cost-based implementation as discussed in section~\ref{sec:evaluation:success_cost_based}.

\section{Related Work}
\label{sec:related_work}

There are several implementations of the ACT-R theory in different programming languages. First of all, there is the official ACT-R implementation in Lisp \cite{actr_homepage} which we used as a reference. There are a lot of extensions to this implementation which partly have been included to the original package in later versions like the ACT-R/PM extension included in ACT-R 6.0 \cite[p. 264]{actr_reference}. The implementation comes with an experiment environment offering a graphical user interface to load, execute and observe models.\looseness=-1

In \cite{stewart_deconstructing_2006,stewart_deconstructing_2007}, a Python implementation is presented which also has the aim to simplify and harmonize parts of the ACT-R theory by finding the central components of the theory. The architecture has been reduced to only the procedural and the declarative memory which are used to build other models combining and adapting them in different ways. However, there is no possibility to translate traditional ACT-R models automatically to Python code since the way of modeling differs too much from the original implementation. 

Furthermore, there are two different implementations in Java: \emph{jACT-R} \cite{jactr} and \emph{ACT-R: The Java Simulation \& Development Environment} \cite{java_actr}. The latter one is capable of executing original ACT-R models and offers an advanced graphical user interface. The focus of the project was to make ACT-R more portable with the help of Java \cite{java_actr_benefits}. In jACT-R, the focus was to offer a clean and exchangeable interface to all the components, so different versions of the ACT-R theory can be mixed \cite{jactr_benefits} and models are defined using XML. There is no compiler from original ACT-R models to XML models of jACT-R. Due to the modular design defining various interfaces which can be exchanged, jACT-R is highly adaptable to personal needs. However, both approaches are missing the proximity to a formal representation.


\section{Conclusion}
\label{sec:conclusion}

In this work, we have presented an implementation of ACT-R using Constraint Handling Rules which is capable of closing the gap between the theory of ACT-R and its technical realization. Our implementation abstracts from technical artifacts and is near to the theory but can reproduce the results of the reference implementation. Furthermore, the formalization itself enables implementations to check against this reference. The implementation of the different conflict resolution strategies has shown the adaptability of our approach. Most of the implemented strategies are not available for the current implementation of ACT-R and our implementation of production rule refraction is unique.

For the future, the implementation can be extended by other modules like the perceptive/motor modules provided by ACT-R. Currently, there is a running student project on implementing a temporal module which may be used to investigate time perception. The formalization and CHR translation pave the way to develop analysis tools (e.g. a confluence test) on the basis of the results for CHR programs.

\newpage


\bibliography{bibliography}

\newpage

\appendix

\section{Formalization of ACT-R}
\label{sec:formalization}

In this section, the fundamental concepts of ACT-R are formalized and transferred to CHR.

\subsection{The Basic Unit of Knowledge: Chunks}
\label{sec:basic_unit_of_knowledge}

ACT-R is a \emph{symbolic} production rule system, i.e. all declarative information is represented in form of symbols and associations of symbols and the procedural information is stored in form of production rules transforming the declarative information. Hence, the ACT-R production system is defined over a set of symbols $\mathfrak{S}$. The smallest unit of declarative information is a \emph{chunk}, which basically is a structured assembly of symbols. It has a unique name and a number of labeled \emph{slots} which can hold one single \emph{symbol}. The chunk names and the slot labels are symbols themselves. If a chunk has a symbol naming a chunk in its slot, the two chunks are \emph{connected}. We require the \emph{unique-name assumption} for symbols. The concept of chunks and their connections in form of chunk stores is defined in section~\ref{sec:chunk_stores}. 

\subsection{Chunk Stores}
\label{sec:chunk_stores}

We extend the abstract notion of chunks given in section~\ref{sec:basic_unit_of_knowledge} to a definition of chunk descriptions embedded into chunk stores which represent a network of chunks with the help of three relations.

\begin{definition}[Chunk Description]
\label{def:chunk_description}
A chunk with name $c$ and type $t$ and corresponding slots and values can be represented as a term $chunk(c,t,\{ (s,v) \enspace | \enspace c \enspace \mathrm{has \enspace the \enspace value} \enspace v \enspace \mathrm{in \enspace slot} \enspace s \})$.
\end{definition}

\begin{definition}[Chunk Store]
\label{def:chunk_store}
A \emph{chunk-store} $\Gamma$ over a set of symbols $\mathfrak{S}$ is a tuple $(\mathbb{C},\mathbb{E},\mathcal{T},\rel{HasSlot},\rel{Isa})$, where $\mathcal{C}$ is a set of chunk identifiers and $\mathbb{E}$ a set of primitive elements both identified by unique names.  The \emph{values} of $\Gamma$ are defined by the set $\mathbb{V} = \mathbb{C} \cup \mathbb{E}$. $\mathcal{T}$ is a set of chunk-types. The set $\mathbb{T}$ then denotes the set of all type names. A chunk-type $T = (t,S) \in \mathcal{T}$ is a tuple with a \emph{unique} type name $t \in \mathbb{T}$ and a set of slots $S \subseteq \mathbb{S}$ where $\mathbb{S}$ is the set of all slot names. The sets $\mathbb{C}$, $\mathbb{E}$, $\mathbb{T}$ and $\mathbb{S}$ are disjoint: $\mathbb{C} \enspace \dot\cup \enspace \mathbb{E} \enspace \dot\cup \enspace \mathbb{T} \enspace \dot\cup \enspace \mathbb{S} \subseteq \mathfrak{S}$.

$\rel{HasSlot} \subseteq \mathbb{C} \times \mathbb{S} \times \mathbb{V}$ and $\rel{Isa} \subseteq \mathbb{C} \times \mathbb{T}$ are relations and are defined as follows: 
\begin{equation*}
chunk(c,t,S) \in \Gamma  \Leftrightarrow c \in \mathbb{C}  \wedge  c \enspace \rel{Isa} \enspace t \wedge \forall (s,v) \in S: (c,s,v) \in \rel{HasSlot}.
\end{equation*}
The $\rel{Isa}$ relation has to be right-unique and left-total, so each chunk has to have exactly one type. A chunk-store is \emph{type-consistent}, iff the following two conditions hold:
\begin{enumerate}
 \item $\forall c \in \mathbb{C}, s \in \mathbb{S}, v,v' \in \mathbb{V}: (c,s,v) \in \rel{HasSlot} \wedge (c,s,v') \in \rel{HasSlot} \Rightarrow v = v'$
 \item $\forall c \in \mathbb{C}, s \in \mathbb{S}, v \in \mathbb{V}, t \in \mathbb{T}, S \subseteq \mathbb{S}: c \enspace \rel{Isa} \enspace t \wedge (c,s,v) \in \rel{HasSlot} \wedge (t,S) \in \mathcal{T} \Rightarrow s \in S$
\end{enumerate}




\end{definition}

With this definition, a chunk store can be implemented directly in CHR by defining the sets and relations as constraints. The constraint \verb|chunk(C,T)| is a condensed representation of the set of chunk symbols $C$ and the $\rel{Isa}$ relation. This is possible, since each chunk in $\mathbb{C}$ has exactly one type. The ternary $\rel{HasSlot}$ relation is represented by constraints of the form \verb|chunk_has_slot(C,S,V)| stating that $(c,s,v) \in \rel{HasSlot}$.

Chunk types can be represented by a constraint \verb|chunk_type(T,S)| where \verb|T| is the symbol denoting the chunk type and \verb|S| is a list of symbols for the slots. Note that there can be added rules to ensure type-consistency and uniqueness of the relations as defined in definition~\ref{def:chunk_store}.

\subsection{Buffer Systems}
\label{sec:buffer_systems}

\begin{definition}[buffer system]
\label{def:buffer_system}
A \emph{buffer system} is a tuple $(B,\Gamma,\rel{Holds})$, where $B$ is a set of buffer names, $\Gamma = (\mathbb{C}, \allowbreak \mathbb{E}, \allowbreak \mathcal{T}, \allowbreak \rel{HasSlot}, \allowbreak \rel{Isa})$ a type-consistent chunk-store and $\rel{Holds} \subseteq B \times C$ a right-unique relation that assigns every buffer at most one chunk that it holds. Buffers which do not appear in the $\rel{Holds}$ relation are called \emph{empty}.

A buffer system is \emph{consistent}, if every chunk that appears in $\rel{Holds}$ is a member of $C$ and $\Sigma$ is a type-consistent chunk-store. It is \emph{clean}, if its chunk-store only holds chunks which appear in $\rel{Holds}$.
\end{definition}


In CHR, the set $B$ and the $\rel{Holds}$ relation can be represented as a constraint \verb|buffer/3| which holds the name of the buffer, the corresponding module (needed for requests) and the name of the chunk it holds as a reference to the chunk store. This is possible since each buffer holds at most one chunk. Empty buffers can be represented by the empty symbol \verb|nil|. For each buffer, there must be exactly one \verb|buffer| constraint. This transforms the $\rel{Holds}$ relation to a left-total and right-unique relation.

\subsubsection{Production Rules}
\label{sec:production_rules}

\begin{definition}[Production Rules]
\label{def:production_rules}
An ACT-R production rule is of the form \verb|(p name buffer_test* ==> action*)| where \verb|name| is a unique symbol indicating the name of the rule. Buffer tests are also denoted as the left-hand-side (LHS), actions as the right-hand-side (RHS) of a rule. A buffer test has the form \verb|=buffer> isa t  s1 v1 ... sn vn| where the symbol \verb|buffer| references the name of the tested buffer and the rest stands for a chunk description $chunk(c,\mathtt{t},\{ (\mathtt{s_1}, \mathtt{v_1}), \dots , (\mathtt{s_n},\mathtt{v_n}) \})$ for a chunk with arbitrary name $c$. The values $\mathtt{v_i}$ can be symbols or variable symbols, where variable symbols are indicated by the prefix \verb|=|. 

An action has the form \verb|#buffer> s1 v1 ... sn vn| where the \verb|#| is a place-holder for the available actions \verb|=|, \verb|+|, \verb|-| denoting modifications, requests and clearings respectively. The other symbols are defined as for the buffer tests. Note that for requests, the first slot symbol must be \verb|isa| followed by a chunk type as value. The values might be variables again, but have to be bound on the left-hand-side of the rule i.e. appear on LHS.
\end{definition}

\begin{definition}[Applicability of a Production Rule]
\label{def:applicability_of_production_rules}
A production rule with buffer tests \verb|=b|$\mathtt{_1}$\verb|> isa| $\mathtt{\enspace t_1 \enspace s_{1,1} \enspace v_{1,1} \enspace \dots s_{1,n_1} \enspace v_{1,n_1} \enspace \dots \enspace}$ \verb|=b| $\mathtt{_k}$\verb|> isa| $\mathtt{\enspace t_k \enspace \dots \enspace s_{k,n_k} \enspace v_{k,n_k}}$ is applicable in a buffer system iff $\exists \bar{v} \in \mathbb{V}: b_1 \in B \wedge \exists c_1 \in C  \wedge  c_1 \enspace \rel{Isa} \enspace t \wedge (c_1,s_{1,1},v_{1,1}) \in \rel{HasSlot} \dots \wedge \bar{v} = \bar{x}$ where $\bar{v}$ denotes a set of values in $\mathbb{V}$ and $\bar{x}$ the variable symbols used on the LHS.
\end{definition}

\subsubsection{Translation of rules}

The production rules as defined in definition~\ref{def:production_rules} operate on the buffer system: They match the content of the buffers and transform it with a defined set of actions. Hence, an ACT-R rule can be transferred to a CHR rule \verb!H ==> G | B!, where the head \verb|H| and guard \verb|G| represent the applicability condition of the rule as defined in definition~\ref{def:applicability_of_production_rules} and the body \verb|B| contains the actions.

\paragraph{Applicability Condition}

The applicability of a rule in definition~\ref{def:applicability_of_production_rules} can be translated directly to the CHR counterparts of the relations. I.e. each relational condition in the applicability condition is expressed by the respective constraint in the head of the rule. The guard is filled with the conditions from $\bar{v} = \bar{x}$. Note that the condition has a set-based semantics (since idempotency can be reduced in classical logic). I.e., for the special case of duplicate tests on the LHS of a production rule, additional rules have to be generated with all possible combinations of unifications of duplicate pairs to implement the set-based semantics of CHR $\omega_\mathit{set}$ as shown in~\cite{Sarna-starosta07compilingconstraint}. In the following, we assume that the production rules are duplicate-free.


\paragraph{Actions} 

The actions of a production rule transform the buffer system in the way as they have been defined in section~\ref{sec:modular_architecture}. The transformations of the buffer system can be realized in CHR by using destructive update as described in \cite[p. 32]{fru_chr_book_2009}. I.e. each action has a trigger constraint \verb|action/2| which gets the name of the buffer and the specification of the action encoded as a chunk-description (see definitions~\ref{def:chunk_description}~and~\ref{def:production_rules}). The trigger constraints then use abstract methods to access the buffer system like \verb|set_buffer| to set the content of a buffer. This simplifies the compilation and the form of the resulting rules, since the constraints representing the relations of the buffer system only appear in the kept head of the resulting CHR rules and never in the removed head. Additionally, it simplifies extensions and adaptations of the actions, since the compiler must not be changed but only the framework implementing the actions. One adaptation of the simplest form of actions which only apply the changes to the buffers is shown in section~\ref{sec:timing} when we introduce scheduling to postpone the actual application of the changes an action performs.

\section{Evaluation Results}
\label{sec:evaluation_results}

In this appendix we list the results of our experiments as described in section~\ref{sec:evaluation}.

\subsection{Samples}

In table~\ref{tab:samples1}~and~\ref{tab:samples2}, the used samples of player~2 and player~3 are listed. Table~\ref{tab:samples1_freq}~and~\ref{tab:samples2_freq} show the frequencies of rock, paper and scissors within one sample. The sum $\Sigma$ is a control value and ensures that 20~moves have been produced per sample. The $p$ values denote the probabilities of rock, paper and scissors respectively.


\begin{table}[ht]
\caption{Samples for Player 2}
\label{tab:samples1}
\begin{tabular}{rrrrrrrrrrrrrrrrrrrrr}
\toprule
Sample  & &&&&&&&&&&&&&&&&&&& \\
\midrule
1  & r & r & r & p & r & p & p & p & r & r & r & r & p & r & p & p & p & r & p & r \\
2  & r & p & p & p & p & p & r & r & p & r & r & r & r & p & r & p & r & r & p & p \\
3  & r & p & r & p & r & r & p & p & r & p & r & p & r & p & r & r & r & r & p & p \\
4  & r & r & r & r & r & r & p & r & p & r & r & p & p & p & p & p & r & r & p & p \\
5  & p & p & p & p & p & r & p & p & r & p & p & r & p & r & r & p & p & r & p & p \\
6  & p & p & p & p & r & p & r & r & r & r & r & r & r & r & r & p & p & p & r & p \\
7  & p & p & r & p & p & p & p & r & r & r & r & r & r & p & r & r & r & r & r & r \\
8  & r & r & p & r & r & p & r & p & p & r & r & r & r & p & r & r & r & p & r & r \\
9  & p & p & p & p & p & p & r & r & p & p & r & r & r & r & r & r & p & r & r & r \\
10 & p & p & p & p & p & p & p & p & p & r & r & p & r & p & p & r & r & p & p & p \\
11 & p & r & p & p & r & p & r & r & r & r & p & p & p & r & r & p & r & p & p & p \\
12 & p & p & r & r & p & r & r & p & r & r & r & p & p & p & r & p & r & r & r & p \\
13 & p & r & p & r & p & r & r & p & r & p & r & r & p & r & p & r & p & p & r & p \\
14 & p & r & p & p & r & r & p & p & r & p & r & p & p & r & p & p & r & p & p & p \\
15 & p & r & p & p & p & p & p & r & r & r & r & r & r & r & r & p & r & p & p & r \\
16 & r & r & p & r & p & p & p & r & p & r & r & r & r & r & p & r & p & r & r & p \\
17 & r & r & r & r & r & p & p & p & r & r & r & r & p & p & p & p & r & r & p & p \\
18 & r & r & r & r & r & r & r & p & r & p & r & r & p & r & r & p & p & p & p & r \\
19 & r & p & r & r & r & p & p & p & p & p & r & p & r & p & r & p & r & p & r & p \\
20 & p & p & r & r & p & p & r & r & p & r & r & p & r & r & p & r & r & p & r & p \\
\bottomrule
\end{tabular}
\end{table}

\begin{table}[ht]
\caption{Samples for Player 2 -- Frequencies and Probabilities}
\label{tab:samples1_freq}
\begin{tabular}{rrrrrrrr}
\toprule
Sample  & \#rock & \#paper & \#scissors & $\Sigma$ & $p_r$   & $p_p$   & $p_s$ \\
\midrule
1       & 11     & 9       & 0          & 20 & 0.55   & 0.45   & 0    \\
2       & 10     & 10      & 0          & 20 & 0.5    & 0.5    & 0    \\
3       & 11     & 9       & 0          & 20 & 0.55   & 0.45   & 0    \\
4       & 11     & 9       & 0          & 20 & 0.55   & 0.45   & 0    \\
5       & 6      & 14      & 0          & 20 & 0.3    & 0.7    & 0    \\
6       & 11     & 9       & 0          & 20 & 0.55   & 0.45   & 0    \\
7       & 13     & 7       & 0          & 20 & 0.65   & 0.35   & 0    \\
8       & 14     & 6       & 0          & 20 & 0.7    & 0.3    & 0    \\
9       & 11     & 9       & 0          & 20 & 0.55   & 0.45   & 0    \\
10      & 5      & 15      & 0          & 20 & 0.25   & 0.75   & 0    \\
11      & 9      & 11      & 0          & 20 & 0.45   & 0.55   & 0    \\
12      & 11     & 9       & 0          & 20 & 0.55   & 0.45   & 0    \\
13      & 10     & 10      & 0          & 20 & 0.5    & 0.5    & 0    \\
14      & 7      & 13      & 0          & 20 & 0.35   & 0.65   & 0    \\
15      & 11     & 9       & 0          & 20 & 0.55   & 0.45   & 0    \\
16      & 12     & 8       & 0          & 20 & 0.6    & 0.4    & 0    \\
17      & 11     & 9       & 0          & 20 & 0.55   & 0.45   & 0    \\
18      & 13     & 7       & 0          & 20 & 0.65   & 0.35   & 0    \\
19      & 9      & 11      & 0          & 20 & 0.45   & 0.55   & 0    \\
20      & 11     & 9       & 0          & 20 & 0.55   & 0.45   & 0    \\
\midrule
Average & 10.35  & 9.65    & 0          & 20 & 0.5175 & 0.4825 & 0   \\
\bottomrule
\end{tabular}
\end{table}


\begin{table}[ht]
\caption{Samples for Player 3}
\label{tab:samples2}
\begin{tabular}{rrrrrrrrrrrrrrrrrrrrr}
\toprule
Sample &   &   &   &   &   &   &   &   &   &   &   &   &   &   &   &   &   &   &   &   \\
\midrule
1      & s & s & s & p & s & s & r & s & r & s & r & s & s & s & r & r & p & s & r & p \\
2      & p & r & p & s & p & p & s & r & r & s & s & r & s & s & s & s & s & r & p & s \\
3      & s & p & r & r & p & p & r & s & r & p & r & s & s & s & r & s & r & p & p & s \\
4      & r & s & s & p & r & s & p & p & r & p & s & p & r & r & s & r & r & p & r & p \\
5      & r & p & p & p & s & s & p & r & r & p & s & r & r & p & p & r & s & p & p & s \\
6      & s & s & p & s & s & r & p & r & p & p & p & s & r & p & p & p & p & s & s & r \\
7      & s & p & r & p & s & p & r & r & p & p & p & r & r & s & s & r & r & p & p & p \\
8      & s & s & r & p & r & r & r & p & s & p & r & s & s & p & p & r & p & s & p & p \\
9      & r & r & s & r & r & r & r & s & p & r & r & s & r & p & s & r & r & s & p & r \\
10     & p & r & r & r & r & p & s & p & s & r & p & s & r & r & s & s & s & s & s & s \\
11     & r & r & s & p & s & s & s & s & r & s & s & p & p & p & p & r & s & s & p & s \\
12     & r & s & r & p & r & s & s & r & r & p & r & r & p & p & r & s & r & p & r & r \\
13     & s & r & r & s & r & s & p & s & p & p & p & p & s & r & s & s & p & r & r & p \\
14     & p & p & p & r & r & s & r & s & r & p & p & s & r & s & r & s & r & p & p & p \\
15     & r & s & s & s & r & p & r & s & s & s & r & s & s & p & s & r & s & s & s & s \\
16     & s & r & p & r & s & r & p & r & r & p & r & r & s & r & r & r & r & r & r & r \\
17     & s & p & s & s & p & s & r & p & p & p & r & s & r & s & r & r & s & r & s & p \\
18     & r & r & s & p & r & s & p & p & p & r & r & p & r & r & p & s & r & s & p & s \\
19     & r & r & s & s & r & s & s & s & r & r & s & s & r & p & s & s & r & p & r & p \\
20     & r & s & p & r & r & s & s & r & p & r & p & p & r & p & p & s & s & s & r & r \\
\bottomrule
\end{tabular}
\end{table}

\begin{table}[ht]
\caption{Samples for Player 3 -- Frequencies and Probabilities}
\label{tab:samples2_freq}
\begin{tabular}{rrrrrrrr}
\toprule
Sample  & \#rock & \#paper & \#scissors & $\Sigma$ & $p_r$   & $p_p$   & $p_s$ \\
\midrule
1       & 6    & 3    & 11   & 20 & 0.3  & 0.15 & 0.55 \\
2       & 5    & 5    & 10   & 20 & 0.25 & 0.25 & 0.5  \\
3       & 7    & 6    & 7    & 20 & 0.35 & 0.3  & 0.35 \\
4       & 8    & 7    & 5    & 20 & 0.4  & 0.35 & 0.25 \\
5       & 6    & 9    & 5    & 20 & 0.3  & 0.45 & 0.25 \\
6       & 4    & 9    & 7    & 20 & 0.2  & 0.45 & 0.35 \\
7       & 7    & 9    & 4    & 20 & 0.35 & 0.45 & 0.2  \\
8       & 6    & 8    & 6    & 20 & 0.3  & 0.4  & 0.3  \\
9       & 12   & 3    & 5    & 20 & 0.6  & 0.15 & 0.25 \\
10      & 7    & 4    & 9    & 20 & 0.35 & 0.2  & 0.45 \\
11      & 4    & 6    & 10   & 20 & 0.2  & 0.3  & 0.5  \\
12      & 11   & 5    & 4    & 20 & 0.55 & 0.25 & 0.2  \\
13      & 6    & 7    & 7    & 20 & 0.3  & 0.35 & 0.35 \\
14      & 7    & 8    & 5    & 20 & 0.35 & 0.4  & 0.25 \\
15      & 5    & 2    & 13   & 20 & 0.25 & 0.1  & 0.65 \\
16      & 14   & 3    & 3    & 20 & 0.7  & 0.15 & 0.15 \\
17      & 6    & 6    & 8    & 20 & 0.3  & 0.3  & 0.4  \\
18      & 8    & 7    & 5    & 20 & 0.4  & 0.35 & 0.25 \\
19      & 8    & 3    & 9    & 20 & 0.4  & 0.15 & 0.45 \\
20      & 8    & 6    & 6    & 20 & 0.4  & 0.3  & 0.3  \\
\midrule
Average & 7.25 & 5.80 & 6.95 & 20 & 0.36 & 0.29 & 0.35\\
\bottomrule
\end{tabular}
\end{table}

\FloatBarrier

\subsection{Reinforcement-Learning-Based Utility Learning}
\label{sec:evaluation_results_reinf}

Tables~\ref{tab:actr6_1},~\ref{tab:actr6_2}~and~\ref{tab:actr6_3} show the results of the ACT-R implementation of player~1,~2~and~3 respectively. In tables~\ref{tab:chr6_1},~\ref{tab:chr6_2}~and~\ref{tab:chr6_3} the results of our CHR implementation can be found. The $U$ values denote the utilities for rock, paper and scissors respectively, where the other values show the performance of the model in the corresponding sample as a control of equal program flows of the two implementations.


\begin{table}[ht]
\caption{Results for Player 1 -- Reward-Based Utility Learning (ACT-R)}
\label{tab:actr6_1}
\begin{tabular}{rrrrrrr}
\toprule
\multicolumn{1}{r}{Sample} &
\multicolumn{3}{c}{Utilities}    &
\multicolumn{3}{c}{Performance}    \\ 
\cmidrule(lr){2-4}
\cmidrule(lr){5-7}
       & $U_r$    & $U_p$    & $U_s$    & \#win  & \#draw & \#defeat \\
\midrule
1 & 0.000 & 1.873 & -0.020 & 19    & 0      & 1     \\
\bottomrule
\end{tabular}
\end{table}


\begin{table}[ht]
\caption{Results for Player 1 -- Reward-Based Utility Learning (CHR)}
\label{tab:chr6_1}
\begin{tabular}{rrrrrrr}
\toprule
\multicolumn{1}{r}{Sample} &
\multicolumn{3}{c}{Utilities}    &
\multicolumn{3}{c}{Performance}    \\ 
\cmidrule(lr){2-4}
\cmidrule(lr){5-7}
       & $U_r$    & $U_p$    & $U_s$    & \#win  & \#draw & \#defeat \\
\midrule
1 &  0.000 & 1.873 & -0.020 & 19    & 0      & 1   \\
\bottomrule
\end{tabular}
\end{table}


\begin{table}[ht]
\caption{Results for Player 2 -- Reward-Based Utility Learning (ACT-R)}
\label{tab:actr6_2}
\begin{tabular}{rrrrrrr}
\toprule
\multicolumn{1}{r}{Sample} &
\multicolumn{3}{c}{Utilities}    &
\multicolumn{3}{c}{Performance}    \\ 
\cmidrule(lr){2-4}
\cmidrule(lr){5-7}
       & $U_r$    & $U_p$    & $U_s$    & \#win  & \#draw & \#defeat \\
\midrule
1       & 0.000     & 1.799 & $-0.020$ & 10     & 9      & 1        \\
2       & 0.000     & 1.822 & $-0.020$ & 9      & 10     & 1        \\
3       & 0.000     & 1.833 & $-0.020$ & 10     & 9      & 1        \\
4       & 0.000     & 1.743 & $-0.020$ & 10     & 9      & 1        \\
5       & 0.000     & 0.000 & 1.267 & 14     & 0      & 6        \\
6       & 0.000 & 0.000 & 0.977 & 9      & 0      & 11       \\
7       & 0.000 & 0.000 & 0.084 & 7      & 0      & 13       \\
8       & 0.000 & 1.849 & $-0.020$ & 13     & 6      & 1        \\
9       & 0.000 & 0.000 & 0.248 & 9      & 0      & 11       \\
10      & 0.000 & 0.000 & 1.329 & 15     & 0      & 5        \\
11      & 0.000 & 0.000 & 1.289 & 11     & 0      & 9        \\
12      & 0.000 & 0.000 & 0.775 & 9      & 0      & 11       \\
13      & 0.000 & 0.000 & 1.076 & 10     & 0      & 10       \\
14      & 0.000 & 0.000 & 1.442 & 13     & 0      & 7        \\
15      & 0.000 & 0.000 & 0.721 & 9      & 0      & 11       \\
16      & 0.000 & 1.836 & $-0.020$ & 11     & 8      & 1        \\
17      & 0.000 & 1.763 & $-0.020$ & 10     & 9      & 1        \\
18      & 0.000 & 1.760  & $-0.020$ & 12     & 7      & 1        \\
19      & 0.000 & 1.805 & $-0.020$ & 8      & 11     & 1        \\
20      & 0.000 & 0.000 & 0.834 & 9      & 0      & 11       \\
\midrule
Average & 0.000 & 0.811 & 0.493 & 10.400 & 3.900  & 5.700   \\
\bottomrule
\end{tabular}
\end{table}


\begin{table}[ht]
\caption{Results for Player 2 -- Reward-Based Utility Learning (CHR)}
\label{tab:chr6_2}
\begin{tabular}{rrrrrrr}
\toprule
\multicolumn{1}{r}{Sample} &
\multicolumn{3}{c}{Utilities}    &
\multicolumn{3}{c}{Performance}    \\ 
\cmidrule(lr){2-4}
\cmidrule(lr){5-7}
       & $U_r$    & $U_p$    & $U_s$    & \#win  & \#draw & \#defeat \\
\midrule
1  & 0.000 & 1.799 & $-0.020$ & 10    & 9      & 1        \\
2  & 0.000 & 1.822 & $-0.020$ & 9     & 10     & 1        \\
3  & 0.000 & 1.833 & $-0.020$ & 10    & 9      & 1        \\
4  & 0.000 & 1.743 & $-0.020$ & 10    & 9      & 1        \\
5  & 0.000 & 0.000 & 1.267  & 14    & 0      & 6        \\
6  & 0.000 & 0.000 & 0.977  & 9     & 0      & 11       \\
7  & 0.000 & 0.000 & 0.084  & 7     & 0      & 13       \\
8  & 0.000 & 1.849 & $-0.020$ & 13    & 6      & 1        \\
9  & 0.000 & 0.000 & 0.248  & 9     & 0      & 11       \\
10 & 0.000 & 0.000 & 1.329  & 15    & 0      & 5        \\
11 & 0.000 & 0.000 & 1.289  & 11    & 0      & 9        \\
12 & 0.000 & 0.000 & 0.775  & 9     & 0      & 11       \\
13 & 0.000 & 0.000 & 1.076  & 10    & 0      & 10       \\
14 & 0.000 & 0.000 & 1.442  & 13    & 0      & 7        \\
15 & 0.000 & 0.000 & 0.721  & 9     & 0      & 11       \\
16 & 0.000 & 1.836 & $-0.020$ & 11    & 8      & 1        \\
17 & 0.000 & 1.763 & $-0.020$ & 10    & 9      & 1        \\
18 & 0.000 & 1.760 & $-0.020$ & 12    & 7      & 1        \\
19 & 0.000 & 1.805 & $-0.020$ & 8     & 11     & 1        \\
20 & 0.000 & 0.000 & 0.834  & 9     & 0      & 11      \\
\midrule
Average & 0.000 & 0.811 & 0.493 & 10.400 & 3.900 &5.700 \\
\bottomrule
\end{tabular}
\end{table}


\begin{table}[ht]
\caption{Results for Player 3 -- Reward-Based Utility Learning (ACT-R)}
\label{tab:actr6_3}
\begin{tabular}{rrrrrrr}
\toprule
\multicolumn{1}{r}{Sample} &
\multicolumn{3}{c}{Utilities}    &
\multicolumn{3}{c}{Performance}    \\ 
\cmidrule(lr){2-4}
\cmidrule(lr){5-7}
       & $U_r$    & $U_p$    & $U_s$    & \#win  & \#draw & \#defeat \\
\midrule
1       & 0.000  & 0.533  & $-0.091$ & 3     & 12     & 5        \\
2       & 0.000  & $-0.052$ & $-0.159$ & 4     & 10     & 6        \\
3       & 0.000  & 0.000  & 0.689  & 6     & 7      & 7        \\
4       & 0.257  & $-0.020$ & $-0.020$ & 4     & 7      & 9        \\
5       & 0.720  & $-0.131$ & $-0.020$ & 4     & 8      & 8        \\
6       & 0.000  & 0.000  & 0.633  & 9     & 7      & 4        \\
7       & 0.000  & 0.000  & 1.046  & 9     & 4      & 7        \\
8       & 0.000  & 0.732  & $-0.090$ & 5     & 10     & 5        \\
9       & 0.000  & 1.254  & $-0.020$ & 11    & 3      & 6        \\
10      & 0.000  & 0.000  & 0.363  & 4     & 9      & 7        \\
11      & 0.000  & 0.434  & $-0.020$ & 3     & 6      & 11       \\
12      & $-0.052$ & 1.575  & $-0.075$ & 7     & 6      & 7        \\
13      & 0.274  & $-0.092$ & $-0.052$ & 4     & 9      & 7        \\
14      & 0.000  & 0.000  & 0.977  & 8     & 5      & 7        \\
15      & 1.717  & $-0.020$ & $-0.020$ & 12    & 4      & 4        \\
16      & 0.000  & 1.764  & $-0.052$ & 13    & 4      & 3        \\
17      & 0.000  & 0.000  & 0.736  & 6     & 8      & 6        \\
18      & 0.000  & 0.541  & $-0.020$ & 7     & 7      & 6        \\
19      & 0.000  & 1.083  & $-0.020$ & 7     & 3      & 10       \\
20      & $-0.020$ & 0.918  & $-0.036$ & 6     & 5      & 9        \\
\midrule
Average & 0.145  & 0.426  & 0.187  & 6.600 & 6.700  & 6.700   \\
\bottomrule
\end{tabular}
\end{table}


\begin{table}[ht]
\caption{Results for Player 3 -- Reward-Based Utility Learning (CHR)}
\label{tab:chr6_3}
\begin{tabular}{rrrrrrr}
\toprule
\multicolumn{1}{r}{Sample} &
\multicolumn{3}{c}{Utilities}    &
\multicolumn{3}{c}{Performance}    \\ 
\cmidrule(lr){2-4}
\cmidrule(lr){5-7}
       & $U_r$    & $U_p$    & $U_s$    & \#win  & \#draw & \#defeat \\
\midrule
1  & 0.000  & 0.533  & $-0.091$ & 3     & 12     & 5     \\
2  & 0.000  & $-0.052$ & $-0.159$ & 4     & 10     & 6     \\
3  & 0.000  & 0.000  & 0.689  & 6     & 7      & 7     \\
4  & 0.257  & $-0.020$ & $-0.020$ & 4     & 7      & 9     \\
5  & 0.720  & $-0.131$ & $-0.020$ & 4     & 8      & 8     \\
6  & 0.000  & 0.000  & 0.633  & 9     & 7      & 4     \\
7  & 0.000  & 0.000  & 1.046  & 9     & 4      & 7     \\
8  & 0.000  & 0.732  & $-0.090$ & 5     & 10     & 5     \\
9  & 0.000  & 1.254  & $-0.020$ & 11    & 3      & 6     \\
10 & 0.000  & 0.000  & 0.363  & 4     & 9      & 7     \\
11 & 0.000  & 0.434  & $-0.020$ & 3     & 6      & 11    \\
12 & $-0.052$ & 1.575  & $-0.075$ & 7     & 6      & 7     \\
13 & 0.274  & $-0.092$ & $-0.052$ & 4     & 9      & 7     \\
14 & 0.000  & 0.000  & 0.977  & 8     & 5      & 7     \\
15 & 1.717  & $-0.020$ & $-0.020$ & 12    & 4      & 4     \\
16 & 0.000  & 1.764  & $-0.052$ & 13    & 4      & 3     \\
17 & 0.000  & 0.000  & 0.736  & 6     & 8      & 6     \\
18 & 0.000  & 0.541  & $-0.020$ & 7     & 7      & 6     \\
19 & 0.000  & 1.083  & $-0.020$ & 7     & 3      & 10    \\
20 & $-0.020$ & 0.918  & $-0.036$ & 6     & 5      & 9     \\
\midrule
Average   & 0.145  & 0.426  & 0.187  & 6.600 & 6.700  & 6.700 \\
\bottomrule
\end{tabular}
\end{table}

\FloatBarrier

\subsection{Success-/Cost-Based Utility Learning}
\label{sec:evaluation_results_succ}

Tables~\ref{tab:actr5_1},~\ref{tab:actr5_2}~and~\ref{tab:actr5_3} show the results of the ACT-R implementation of player~1,~2 and~3 respectively. In tables~\ref{tab:chr5_1},~\ref{tab:chr5_2}~and~\ref{tab:chr5_3} the results of our CHR implementation can be found. The meaning of the values corresponds to Appendix~\ref{sec:evaluation_results_reinf}.


\begin{table}[ht]
\caption{Results for Player 1 -- Success-/Cost-Based Utility Learning (ACT-R)}
\label{tab:actr5_1}
\begin{tabular}{rrrrrrr}
\toprule
\multicolumn{1}{r}{Sample} &
\multicolumn{3}{c}{Utilities}    &
\multicolumn{3}{c}{Performance}    \\ 
\cmidrule(lr){2-4}
\cmidrule(lr){5-7}
       & $U_r$    & $U_p$    & $U_s$    & \#win  & \#draw & \#defeat \\
\midrule
1 & 19.95 & 19.95 & 19.95 & 0     & 20     & 0    \\
\bottomrule
\end{tabular}
\end{table}


\begin{table}[ht]
\caption{Results for Player 1 -- Success-/Cost-Based Utility Learning (CHR)}
\label{tab:chr5_1}
\begin{tabular}{rrrrrrr}
\toprule
\multicolumn{1}{r}{Sample} &
\multicolumn{3}{c}{Utilities}    &
\multicolumn{3}{c}{Performance}    \\ 
\cmidrule(lr){2-4}
\cmidrule(lr){5-7}
       & $U_r$    & $U_p$    & $U_s$    & \#win  & \#draw & \#defeat \\
\midrule
1 & 19.95 & 19.95 & 19.95 & 0     & 20     & 0      \\
\bottomrule
\end{tabular}
\end{table}


\begin{table}[ht]
\caption{Results for Player 2 -- Success-/Cost-Based Utility Learning (ACT-R)}
\label{tab:actr5_2}
\begin{tabular}{rrrrrrr}
\toprule
\multicolumn{1}{r}{Sample} &
\multicolumn{3}{c}{Utilities}    &
\multicolumn{3}{c}{Performance}    \\ 
\cmidrule(lr){2-4}
\cmidrule(lr){5-7}
       & $U_r$    & $U_p$    & $U_s$    & \#win  & \#draw & \#defeat \\
\midrule
1       & 3.790 & 19.830 & 13.250 & 8     & 10     & 2        \\
2       & 6.550 & 19.822 & 9.925  & 8     & 10     & 2        \\
3       & 6.550 & 19.863 & 13.250 & 10    & 8      & 2        \\
4       & 2.144 & 19.755 & 13.250 & 5     & 13     & 2        \\
5       & 9.925 & 19.790 & 14.913 & 7     & 11     & 2        \\
6       & 9.925 & 19.832 & 13.250 & 11    & 7      & 2        \\
7       & 9.925 & 19.890 & 16.575 & 16    & 2      & 2        \\
8       & 4.838 & 19.868 & 9.925  & 11    & 7      & 2        \\
9       & 9.925 & 19.803 & 9.925  & 10    & 8      & 2        \\
10      & 9.925 & 19.653 & 9.925  & 4     & 14     & 2        \\
11      & 9.925 & 19.846 & 14.913 & 10    & 8      & 2        \\
12      & 9.925 & 19.847 & 9.925  & 10    & 8      & 2        \\
13      & 9.925 & 19.859 & 14.913 & 11    & 7      & 2        \\
14      & 9.925 & 19.830 & 13.250 & 7     & 11     & 2        \\
15      & 9.925 & 19.825 & 13.250 & 11    & 7      & 2        \\
16      & 4.838 & 19.868 & 14.913 & 11    & 7      & 2        \\
17      & 2.550 & 19.796 & 9.925  & 5     & 13     & 2        \\
18      & 1.817 & 19.805 & 13.250 & 6     & 12     & 2        \\
19      & 6.550 & 19.791 & 9.925  & 7     & 11     & 2        \\
20      & 9.925 & 19.858 & 9.925  & 10    & 8      & 2        \\
\midrule
Average & 7.440 & 19.822 & 12.419 & 8.9   & 9.1    & 2       \\
\bottomrule
\end{tabular}
\end{table}


\begin{table}[ht]
\caption{Results for Player 2 -- Success-/Cost-Based Utility Learning (CHR)}
\label{tab:chr5_2}
\begin{tabular}{rrrrrrr}
\toprule
\multicolumn{1}{r}{Sample} &
\multicolumn{3}{c}{Utilities}    &
\multicolumn{3}{c}{Performance}    \\ 
\cmidrule(lr){2-4}
\cmidrule(lr){5-7}
       & $U_r$    & $U_p$    & $U_s$    & \#win  & \#draw & \#defeat \\
\midrule
1       & 3.790 & 19.830 & 13.250 & 8     & 10     & 2        \\
2       & 6.550 & 19.822 & 9.925  & 8     & 10     & 2        \\
3       & 6.550 & 19.863 & 13.250 & 10    & 8      & 2        \\
4       & 2.144 & 19.755 & 13.250 & 5     & 13     & 2        \\
5       & 9.925 & 19.790 & 14.913 & 7     & 11     & 2        \\
6       & 9.925 & 19.832 & 13.250 & 11    & 7      & 2        \\
7       & 9.925 & 19.890 & 16.575 & 16    & 2      & 2        \\
8       & 4.838 & 19.868 & 9.925  & 11    & 7      & 2        \\
9       & 9.925 & 19.803 & 9.925  & 10    & 8      & 2        \\
10      & 9.925 & 19.653 & 9.925  & 4     & 14     & 2        \\
11      & 9.925 & 19.846 & 14.913 & 10    & 8      & 2        \\
12      & 9.925 & 19.847 & 9.925  & 10    & 8      & 2        \\
13      & 9.925 & 19.859 & 14.913 & 11    & 7      & 2        \\
14      & 9.925 & 19.830 & 13.250 & 7     & 11     & 2        \\
15      & 9.925 & 19.825 & 13.250 & 11    & 7      & 2        \\
16      & 4.838 & 19.868 & 14.913 & 11    & 7      & 2        \\
17      & 2.550 & 19.796 & 9.925  & 5     & 13     & 2        \\
18      & 1.817 & 19.805 & 13.250 & 6     & 12     & 2        \\
19      & 6.550 & 19.791 & 9.925  & 7     & 11     & 2        \\
20      & 9.925 & 19.858 & 9.925  & 10    & 8      & 2        \\
\midrule
Average & 7.440 & 19.822 & 12.419 & 8.9   & 9.1    & 2       \\
\bottomrule
\end{tabular}
\end{table}


\begin{table}[ht]
\caption{Results for Player 3 -- Success-/Cost-Based Utility Learning (ACT-R)}
\label{tab:actr5_3}
\begin{tabular}{rrrrrrr}
\toprule
\multicolumn{1}{r}{Sample} &
\multicolumn{3}{c}{Utilities}    &
\multicolumn{3}{c}{Performance}    \\ 
\cmidrule(lr){2-4}
\cmidrule(lr){5-7}
       & $U_r$    & $U_p$    & $U_s$    & \#win  & \#draw & \#defeat \\
\midrule
1       & 13.985 & 9.925  & 14.887 & 10    & 6      & 4        \\
2       & 13.228 & 9.908  & 9.881  & 10    & 3      & 7        \\
3       & 11.307 & 11.304 & 9.925  & 6     & 6      & 8        \\
4       & 9.268  & 9.925  & 9.913  & 5     & 6      & 9        \\
5       & 6.550  & 4.838  & 11.883 & 6     & 7      & 7        \\
6       & 7.381  & 3.850  & 9.865  & 6     & 7      & 7        \\
7       & 9.858  & 16.558 & 12.394 & 9     & 7      & 4        \\
8       & 5.536  & 6.583  & 15.258 & 7     & 6      & 7        \\
9       & 11.279 & 14.182 & 9.925  & 9     & 4      & 7        \\
10      & 4.888  & 6.557  & 4.888  & 4     & 4      & 12       \\
11      & 13.210 & 6.550  & 3.790  & 6     & 7      & 7        \\
12      & 13.208 & 19.871 & 7.850  & 11    & 7      & 2        \\
13      & 9.875  & 8.450  & 9.925  & 7     & 5      & 8        \\
14      & 9.925  & 11.116 & 9.925  & 6     & 7      & 7        \\
15      & 17.658 & 9.925  & 6.550  & 11    & 5      & 4        \\
16      & 19.888 & 17.394 & 14.913 & 9     & 9      & 2        \\
17      & 13.875 & 6.550  & 9.875  & 7     & 7      & 6        \\
18      & 17.356 & 16.214 & 14.887 & 8     & 9      & 3        \\
19      & 13.539 & 9.925  & 9.925  & 8     & 7      & 5        \\
20      & 15.439 & 13.835 & 9.925  & 8     & 9      & 3        \\
\midrule
Average & 11.863 & 10.673 & 10.319 & 7.65  & 6.4    & 5.95    \\
\bottomrule
\end{tabular}
\end{table}


\begin{table}[ht]
\caption{Results for Player 3 -- Success-/Cost-Based Utility Learning (CHR)}
\label{tab:chr5_3}
\begin{tabular}{rrrrrrr}
\toprule
\multicolumn{1}{r}{Sample} &
\multicolumn{3}{c}{Utilities}    &
\multicolumn{3}{c}{Performance}    \\ 
\cmidrule(lr){2-4}
\cmidrule(lr){5-7}
       & $U_r$    & $U_p$    & $U_s$    & \#win  & \#draw & \#defeat \\
\midrule
1       & 13.985 & 9.925  & 14.888 & 10    & 6      & 4        \\
2       & 13.228 & 9.908  & 9.881  & 10    & 3      & 7        \\
3       & 11.307 & 11.304 & 9.925  & 6     & 6      & 8        \\
4       & 9.268  & 9.925  & 9.913  & 5     & 6      & 9        \\
5       & 6.550  & 4.838  & 11.883 & 6     & 7      & 7        \\
6       & 7.381  & 3.850  & 9.865  & 6     & 7      & 7        \\
7       & 9.858  & 16.558 & 12.394 & 9     & 7      & 4        \\
8       & 5.536  & 6.583  & 15.258 & 7     & 6      & 7        \\
9       & 11.279 & 14.182 & 9.925  & 9     & 4      & 7        \\
10      & 4.888  & 6.557  & 4.888  & 4     & 4      & 12       \\
11      & 13.210 & 6.550  & 3.790  & 6     & 7      & 7        \\
12      & 13.208 & 19.871 & 7.850  & 11    & 7      & 2        \\
13      & 9.875  & 8.450  & 9.925  & 7     & 5      & 8        \\
14      & 9.925  & 11.116 & 9.925  & 6     & 7      & 7        \\
15      & 17.658 & 9.925  & 6.550  & 11    & 5      & 4        \\
16      & 19.888 & 17.394 & 14.913 & 9     & 9      & 2        \\
17      & 13.875 & 6.550  & 9.875  & 7     & 7      & 6        \\
18      & 17.356 & 16.214 & 14.888 & 8     & 9      & 3        \\
19      & 13.539 & 9.925  & 9.925  & 8     & 7      & 5        \\
20      & 15.439 & 13.835 & 9.925  & 8     & 9      & 3        \\
\midrule
Average & 11.863 & 10.673 & 10.319 & 7.65  & 6.4    & 5.95    \\
\bottomrule
\end{tabular}
\end{table}

\FloatBarrier

\subsection{Random Estimated Costs}
\label{sec:evaluation_results_rand}

Tables~\ref{tab:actro_1},~\ref{tab:actro_2}~and~\ref{tab:actro_3} show the results of the ACT-R implementation of player~1,~2 and~3 respectively. In tables~\ref{tab:chro_1},~\ref{tab:chro_2}~and~\ref{tab:chro_3} the results of our CHR implementation can be found. The results have been produced by the first sample of each player and have been run 50 times. The meaning of the values corresponds to Appendix~\ref{sec:evaluation_results_reinf}. In the tables containing the results of the CHR implementation, we added the error squares of the averages over all runs to compare them to the reference implementation.


\begin{table}[ht]
\caption{Results for Player 1 -- Random Estimated Costs (ACT-R)}
\label{tab:actro_1}
\begin{tabular}{rrrrrrr}
\toprule
\multicolumn{1}{r}{Run} &
\multicolumn{3}{c}{Utilities}    &
\multicolumn{3}{c}{Performance}    \\ 
\cmidrule(lr){2-4}
\cmidrule(lr){5-7}
       & $U_r$    & $U_p$    & $U_s$    & \#win  & \#draw & \#defeat \\
\midrule
1   & 19.970 & 19.942 & 9.937      & 12     & 7        & 1 \\
2   & 19.975 & 19.680 & 9.761      & 11     & 8        & 1 \\
3   & 9.565  & 19.987 & 9.912      & 18     & 1        & 1 \\
4   & 9.836  & 19.813 & 9.900      & 18     & 1        & 1 \\
5   & 19.891 & 19.958 & 9.972      & 14     & 5        & 1 \\
6   & 19.390 & 19.999 & 9.596      & 13     & 6        & 1 \\
7   & 4.918  & 19.874 & 9.940      & 16     & 3        & 1 \\
8   & 19.877 & 19.831 & 9.985      & 11     & 8        & 1 \\
9   & 19.854 & 19.873 & 9.850      & 13     & 6        & 1 \\
10  & 19.802 & 19.825 & 9.918      & 11     & 8        & 1 \\
11  & 3.630  & 19.997 & 9.837      & 16     & 3        & 1 \\
12  & 19.995 & 19.985 & 9.878      & 11     & 8        & 1 \\
13  & 6.403  & 19.731 & 9.986      & 17     & 2        & 1 \\
14  & 19.518 & 19.992 & 9.960      & 9      & 10       & 1 \\
15  & 9.958  & 19.958 & 9.994      & 18     & 1        & 1 \\
16  & 19.215 & 19.883 & 9.960      & 9      & 10       & 1 \\
17  & 19.876 & 19.800 & 9.833      & 14     & 5        & 1 \\
18  & 19.992 & 19.900 & 9.854      & 9      & 10       & 1 \\
19  & 19.569 & 19.834 & 9.769      & 15     & 4        & 1 \\
20  & 19.837 & 19.802 & 9.981      & 10     & 9        & 1 \\
21  & 9.971  & 19.950 & 9.500      & 18     & 1        & 1 \\
22  & 19.968 & 19.847 & 9.912      & 11     & 8        & 1 \\
23  & 19.628 & 19.833 & 9.900      & 13     & 6        & 1 \\
24  & 9.887  & 19.936 & 9.934      & 18     & 1        & 1 \\
25  & 9.104  & 19.982 & 9.998      & 18     & 1        & 1 \\
26  & 19.698 & 19.742 & 9.940      & 12     & 7        & 1 \\
27  & 19.665 & 19.992 & 9.920      & 14     & 5        & 1 \\
28  & 13.036 & 19.990 & 9.909      & 17     & 2        & 1 \\
29  & 19.648 & 19.904 & 9.872      & 14     & 5        & 1 \\
30  & 19.994 & 19.898 & 9.837      & 12     & 7        & 1 \\
31  & 19.967 & 19.862 & 9.997      & 11     & 8        & 1 \\
32  & 19.480 & 19.922 & 9.986      & 13     & 6        & 1 \\
33  & 19.978 & 19.883 & 9.839      & 10     & 9        & 1 \\
34  & 19.929 & 19.974 & 9.994      & 14     & 5        & 1 \\
35  & 19.730 & 19.984 & 9.738      & 11     & 8        & 1 \\
36  & 19.511 & 19.974 & 9.886      & 13     & 6        & 1 \\
37  & 9.995  & 19.892 & 9.843      & 18     & 1        & 1 \\
38  & 19.596 & 19.907 & 9.569      & 13     & 6        & 1 \\
39  & 19.837 & 19.802 & 9.981      & 10     & 9        & 1 \\
40  & 9.166  & 19.959 & 9.922      & 18     & 1        & 1 \\
41  & 19.513 & 19.989 & 9.980      & 9      & 10       & 1 \\
42  & 19.669 & 19.943 & 9.711      & 15     & 4        & 1 \\
43  & 3.362  & 19.936 & 9.901      & 15     & 4        & 1 \\
44  & 19.964 & 19.956 & 9.998      & 14     & 5        & 1 \\
45  & 3.761  & 19.653 & 9.596      & 16     & 3        & 1 \\
46  & 9.559  & 19.961 & 9.987      & 18     & 1        & 1 \\
47  & 5.939  & 19.948 & 9.985      & 17     & 2        & 1 \\
48  & 19.632 & 19.942 & 9.724      & 11     & 8        & 1 \\
49  & 19.531 & 19.948 & 9.996      & 9      & 10       & 1 \\
50  & 19.748 & 19.895 & 9.976      & 14     & 5        & 1 \\
\midrule
Average    & 15.991 & 19.901 & 9.883      & 13.62  & 5.38     & 1\\
\bottomrule
\end{tabular}
\end{table}


\begin{table}[ht]
\caption{Results for Player 1 -- Random Estimated Costs (CHR)}
\label{tab:chro_1}
\begin{tabular}{rrrrrrr}
\toprule
\multicolumn{1}{r}{Run} &
\multicolumn{3}{c}{Utilities}    &
\multicolumn{3}{c}{Performance}    \\ 
\cmidrule(lr){2-4}
\cmidrule(lr){5-7}
       & $U_r$    & $U_p$    & $U_s$    & \#win  & \#draw & \#defeat \\
\midrule
1     & 19.645 & 19.969 & 9.986  & 11    & 8      & 1        \\
2     & 19.220 & 19.948 & 9.749  & 10    & 9      & 1        \\
3     & 9.990  & 19.860 & 9.920  & 18    & 1      & 1        \\
4     & 19.838 & 19.941 & 9.922  & 12    & 7      & 1        \\
5     & 19.917 & 19.945 & 9.949  & 12    & 7      & 1        \\
6     & 19.948 & 19.886 & 9.964  & 12    & 7      & 1        \\
7     & 19.997 & 19.777 & 9.903  & 12    & 7      & 1        \\
8     & 19.868 & 19.966 & 9.930  & 10    & 9      & 1        \\
9     & 19.274 & 19.968 & 9.874  & 13    & 6      & 1        \\
10    & 9.810  & 19.937 & 9.975  & 18    & 1      & 1        \\
11    & 19.596 & 19.655 & 9.953  & 10    & 9      & 1        \\
12    & 19.998 & 19.807 & 9.937  & 12    & 7      & 1        \\
13    & 9.868  & 19.858 & 9.600  & 18    & 1      & 1        \\
14    & 19.988 & 19.884 & 9.804  & 13    & 6      & 1        \\
15    & 9.865  & 19.992 & 9.763  & 18    & 1      & 1        \\
16    & 6.509  & 19.965 & 9.947  & 17    & 2      & 1        \\
17    & 19.729 & 19.991 & 9.914  & 14    & 5      & 1        \\
18    & 19.965 & 19.929 & 9.876  & 14    & 5      & 1        \\
19    & 19.941 & 19.907 & 9.985  & 15    & 4      & 1        \\
20    & 19.567 & 19.997 & 9.970  & 16    & 3      & 1        \\
21    & 6.639  & 19.983 & 9.998  & 17    & 2      & 1        \\
22    & 18.898 & 19.760 & 9.953  & 13    & 6      & 1        \\
23    & 19.998 & 19.989 & 9.806  & 8     & 11     & 1        \\
24    & 9.820  & 19.804 & 9.763  & 18    & 1      & 1        \\
25    & 19.968 & 19.684 & 9.937  & 16    & 3      & 1        \\
26    & 19.664 & 19.912 & 9.420  & 12    & 7      & 1        \\
27    & 18.650 & 19.957 & 9.883  & 11    & 8      & 1        \\
28    & 9.754  & 19.938 & 9.711  & 18    & 1      & 1        \\
29    & 19.912 & 19.963 & 10.000 & 10    & 9      & 1        \\
30    & 15.796 & 19.983 & 9.875  & 15    & 4      & 1        \\
31    & 19.100 & 19.951 & 9.998  & 10    & 9      & 1        \\
32    & 9.876  & 19.998 & 9.915  & 18    & 1      & 1        \\
33    & 19.923 & 19.755 & 9.696  & 12    & 7      & 1        \\
34    & 19.779 & 19.739 & 9.948  & 13    & 6      & 1        \\
35    & 19.877 & 19.878 & 9.906  & 10    & 9      & 1        \\
36    & 15.719 & 19.979 & 9.969  & 15    & 4      & 1        \\
37    & 13.243 & 19.882 & 9.842  & 17    & 2      & 1        \\
38    & 19.319 & 19.984 & 9.608  & 12    & 7      & 1        \\
39    & 9.906  & 19.948 & 9.915  & 18    & 1      & 1        \\
40    & 3.165  & 19.862 & 9.960  & 16    & 3      & 1        \\
41    & 19.904 & 19.952 & 9.977  & 13    & 6      & 1        \\
42    & 5.120  & 19.970 & 9.794  & 17    & 2      & 1        \\
43    & 3.746  & 19.787 & 9.855  & 16    & 3      & 1        \\
44    & 19.440 & 19.820 & 9.921  & 11    & 8      & 1        \\
45    & 19.893 & 19.755 & 9.706  & 9     & 10     & 1        \\
46    & 19.953 & 19.874 & 9.797  & 14    & 5      & 1        \\
47    & 19.871 & 19.992 & 9.920  & 12    & 7      & 1        \\
48    & 6.265  & 19.844 & 9.763  & 17    & 2      & 1        \\
49    & 4.798  & 19.823 & 9.868  & 16    & 3      & 1        \\
50    & 9.958  & 19.702 & 9.870  & 18    & 1      & 1        \\
\midrule
Average   & 15.610 & 19.893 & 9.870  & 13.94 & 5.06   & 1        \\
Error Square of Average & 0.145  & 0.000  & 0.000  & 0.10  & 0.10   & 0.00   \\
\bottomrule
\end{tabular}
\end{table}


\begin{table}[ht]
\caption{Results for Player 2 -- Random Estimated Costs (ACT-R)}
\label{tab:actro_2}
\begin{tabular}{rrrrrrr}
\toprule
\multicolumn{1}{r}{Run} &
\multicolumn{3}{c}{Utilities}    &
\multicolumn{3}{c}{Performance}    \\ 
\cmidrule(lr){2-4}
\cmidrule(lr){5-7}
       & $U_r$    & $U_p$    & $U_s$    & \#win  & \#draw & \#defeat \\
\midrule
1  & 2.851  & 14.046 & 6.389  & 6     & 12     & 2        \\
2  & 15.635 & 17.997 & 16.560 & 11    & 7      & 2        \\
3  & 9.429  & 18.123 & 9.769  & 8     & 10     & 2        \\
4  & 13.311 & 19.899 & 9.937  & 9     & 9      & 2        \\
5  & 9.971  & 19.907 & 9.500  & 9     & 10     & 1        \\
6  & 13.230 & 18.308 & 14.984 & 11    & 7      & 2        \\
7  & 12.996 & 19.690 & 9.900  & 9     & 9      & 2        \\
8  & 9.887  & 19.882 & 9.934  & 9     & 10     & 1        \\
9  & 9.104  & 19.966 & 9.998  & 9     & 10     & 1        \\
10 & 4.944  & 19.794 & 14.953 & 10    & 8      & 2        \\
11 & 9.920  & 19.694 & 9.985  & 10    & 8      & 2        \\
12 & 16.936 & 19.769 & 15.888 & 8     & 10     & 2        \\
13 & 6.475  & 19.673 & 9.872  & 9     & 9      & 2        \\
14 & 13.047 & 15.082 & 13.330 & 7     & 11     & 2        \\
15 & 6.659  & 19.743 & 9.976  & 9     & 9      & 2        \\
16 & 12.606 & 19.802 & 9.986  & 6     & 12     & 2        \\
17 & 6.292  & 19.782 & 9.987  & 9     & 9      & 2        \\
18 & 13.257 & 18.705 & 9.994  & 9     & 9      & 2        \\
19 & 7.346  & 17.109 & 9.815  & 8     & 9      & 3        \\
20 & 13.139 & 19.689 & 15.914 & 12    & 6      & 2        \\
21 & 6.327  & 19.813 & 9.997  & 9     & 9      & 2        \\
22 & 3.847  & 18.342 & 9.569  & 8     & 10     & 2        \\
23 & 6.623  & 19.630 & 9.909  & 9     & 9      & 2        \\
24 & 9.166  & 19.924 & 9.922  & 9     & 10     & 1        \\
25 & 4.913  & 19.979 & 9.761  & 8     & 10     & 2        \\
26 & 11.565 & 19.889 & 9.711  & 7     & 11     & 2        \\
27 & 3.538  & 19.892 & 14.923 & 9     & 9      & 2        \\
28 & 7.995  & 19.915 & 13.310 & 8     & 10     & 2        \\
29 & 13.105 & 19.520 & 9.940  & 9     & 9      & 2        \\
30 & 9.735  & 19.985 & 9.920  & 10    & 8      & 2        \\
31 & 13.102 & 19.982 & 9.909  & 9     & 9      & 2        \\
32 & 6.367  & 19.822 & 9.724  & 9     & 9      & 2        \\
33 & 6.657  & 19.903 & 9.841  & 9     & 9      & 2        \\
34 & 4.849  & 15.462 & 6.306  & 7     & 10     & 3        \\
35 & 5.697  & 19.996 & 9.878  & 9     & 9      & 2        \\
36 & 9.259  & 16.904 & 9.916  & 8     & 9      & 3        \\
37 & 9.891  & 19.808 & 9.960  & 9     & 10     & 1        \\
38 & 7.346  & 17.109 & 9.815  & 8     & 9      & 3        \\
39 & 6.400  & 19.952 & 9.689  & 9     & 9      & 2        \\
40 & 9.995  & 19.801 & 9.843  & 9     & 10     & 1        \\
41 & 3.133  & 19.822 & 9.743  & 8     & 10     & 2        \\
42 & 9.833  & 18.376 & 9.688  & 8     & 10     & 2        \\
43 & 3.353  & 17.750 & 8.480  & 6     & 9      & 5        \\
44 & 13.313 & 19.406 & 9.761  & 9     & 9      & 2        \\
45 & 9.565  & 19.976 & 9.912  & 9     & 10     & 1        \\
46 & 9.836  & 19.654 & 9.900  & 9     & 10     & 1        \\
47 & 6.601  & 19.922 & 9.932  & 9     & 9      & 2        \\
48 & 9.364  & 9.125  & 13.513 & 7     & 8      & 5        \\
49 & 9.559  & 19.928 & 9.987  & 9     & 10     & 1        \\
50 & 5.939  & 19.902 & 9.985  & 8     & 11     & 1        \\
\midrule
Average   & 8.878  & 18.923 & 10.588 & 8.64  & 9.36   & 2      \\
\bottomrule
\end{tabular}
\end{table}


\begin{table}[ht]
\caption{Results for Player 2 -- Random Estimated Costs (CHR)}
\label{tab:chro_2}
\begin{tabular}{rrrrrrr}
\toprule
\multicolumn{1}{r}{Run} &
\multicolumn{3}{c}{Utilities}    &
\multicolumn{3}{c}{Performance}    \\ 
\cmidrule(lr){2-4}
\cmidrule(lr){5-7}
       & $U_r$    & $U_p$    & $U_s$    & \#win  & \#draw & \#defeat \\
\midrule
1                        & 13.261 & 19.844 & 13.319 & 10    & 8      & 2        \\
2                        & 6.585  & 17.628 & 9.744  & 9     & 9      & 2        \\
3                        & 16.548 & 17.907 & 17.098 & 11    & 7      & 2        \\
4                        & 9.823  & 19.906 & 9.907  & 10    & 8      & 2        \\
5                        & 14.785 & 19.965 & 9.967  & 8     & 10     & 2        \\
6                        & 11.344 & 11.680 & 9.960  & 5     & 12     & 3        \\
7                        & 15.863 & 19.948 & 9.831  & 7     & 11     & 2        \\
8                        & 9.608  & 19.874 & 9.941  & 9     & 10     & 1        \\
9                        & 9.377  & 19.855 & 9.794  & 9     & 10     & 1        \\
10                       & 11.966 & 19.909 & 15.979 & 9     & 10     & 1        \\
11                       & 4.679  & 19.174 & 9.928  & 8     & 10     & 2        \\
12                       & 8.550  & 16.976 & 9.902  & 6     & 11     & 3        \\
13                       & 9.552  & 19.889 & 9.999  & 9     & 10     & 1        \\
14                       & 9.815  & 19.972 & 9.629  & 9     & 10     & 1        \\
15                       & 9.798  & 19.968 & 9.994  & 9     & 10     & 1        \\
16                       & 9.940  & 19.733 & 9.929  & 10    & 8      & 2        \\
17                       & 9.553  & 19.670 & 9.905  & 9     & 10     & 1        \\
18                       & 13.267 & 19.345 & 9.964  & 9     & 9      & 2        \\
19                       & 9.848  & 19.615 & 9.892  & 10    & 8      & 2        \\
20                       & 6.534  & 19.053 & 9.770  & 9     & 9      & 2        \\
21                       & 6.396  & 16.268 & 9.997  & 4     & 13     & 3        \\
22                       & 19.911 & 19.855 & 9.976  & 5     & 14     & 1        \\
23                       & 9.901  & 19.608 & 9.917  & 9     & 10     & 1        \\
24                       & 14.898 & 18.181 & 15.851 & 11    & 7      & 2        \\
25                       & 13.085 & 18.217 & 9.774  & 9     & 9      & 2        \\
26                       & 6.142  & 19.978 & 9.906  & 8     & 11     & 1        \\
27                       & 9.978  & 19.714 & 9.997  & 9     & 10     & 1        \\
28                       & 9.946  & 19.894 & 9.984  & 10    & 8      & 2        \\
29                       & 13.131 & 18.608 & 9.943  & 9     & 9      & 2        \\
30                       & 6.049  & 19.969 & 9.866  & 9     & 9      & 2        \\
31                       & 9.715  & 19.993 & 9.795  & 10    & 8      & 2        \\
32                       & 6.643  & 19.977 & 9.980  & 9     & 9      & 2        \\
33                       & 12.765 & 19.883 & 9.895  & 9     & 9      & 2        \\
34                       & 9.968  & 19.996 & 9.829  & 10    & 8      & 2        \\
35                       & 4.781  & 17.791 & 9.975  & 8     & 10     & 2        \\
36                       & 4.970  & 17.242 & 9.678  & 8     & 9      & 3        \\
37                       & 6.309  & 19.924 & 9.979  & 9     & 9      & 2        \\
38                       & 14.950 & 19.948 & 9.921  & 8     & 10     & 2        \\
39                       & 8.410  & 19.962 & 9.760  & 9     & 10     & 1        \\
40                       & 4.734  & 12.897 & 9.914  & 3     & 13     & 4        \\
41                       & 9.802  & 19.798 & 9.815  & 9     & 10     & 1        \\
42                       & 3.181  & 16.619 & 6.550  & 5     & 10     & 5        \\
43                       & 9.881  & 19.954 & 9.959  & 9     & 10     & 1        \\
44                       & 6.471  & 19.441 & 9.763  & 8     & 11     & 1        \\
45                       & 9.978  & 18.432 & 9.654  & 8     & 10     & 2        \\
46                       & 9.932  & 19.972 & 9.938  & 9     & 10     & 1        \\
47                       & 7.731  & 17.000 & 9.978  & 6     & 11     & 3        \\
48                       & 6.647  & 18.836 & 9.801  & 9     & 9      & 2        \\
49                       & 13.095 & 19.908 & 9.943  & 9     & 9      & 2        \\
50                       & 9.898  & 17.710 & 9.965  & 10    & 8      & 2        \\
\midrule
Average                  & 9.800  & 18.910 & 10.275 & 8.46  & 9.66   & 1.88     \\
Error Square of Average & 0.850  & 0.000  & 0.098  & 0.032 & 0.090  & 0.014   \\
\bottomrule
\end{tabular}
\end{table}


\begin{table}[ht]
\caption{Results for Player 3 -- Random Estimated Costs (ACT-R)}
\label{tab:actro_3}
\begin{tabular}{rrrrrrr}
\toprule
\multicolumn{1}{r}{Run} &
\multicolumn{3}{c}{Utilities}    &
\multicolumn{3}{c}{Performance}    \\ 
\cmidrule(lr){2-4}
\cmidrule(lr){5-7}
       & $U_r$    & $U_p$    & $U_s$    & \#win  & \#draw & \#defeat \\
\midrule
1       & 15.827 & 9.996  & 9.755  & 10    & 6      & 4        \\
2       & 14.019 & 9.976  & 7.990  & 7     & 8      & 5        \\
3       & 15.242 & 9.986  & 10.701 & 8     & 7      & 5        \\
4       & 13.066 & 6.292  & 9.963  & 6     & 10     & 4        \\
5       & 14.952 & 9.935  & 6.643  & 8     & 8      & 4        \\
6       & 14.255 & 9.815  & 7.259  & 7     & 8      & 5        \\
7       & 14.236 & 9.689  & 7.676  & 7     & 8      & 5        \\
8       & 15.102 & 9.843  & 9.995  & 9     & 7      & 4        \\
9       & 15.215 & 9.569  & 10.978 & 8     & 7      & 5        \\
10      & 13.902 & 9.909  & 7.947  & 7     & 8      & 5        \\
11      & 15.931 & 9.922  & 9.250  & 10    & 6      & 4        \\
12      & 14.264 & 9.761  & 7.943  & 7     & 8      & 5        \\
13      & 15.192 & 9.739  & 9.518  & 9     & 7      & 4        \\
14      & 15.892 & 9.934  & 14.901 & 10    & 6      & 4        \\
15      & 16.847 & 9.998  & 16.631 & 8     & 8      & 4        \\
16      & 14.757 & 9.804  & 10.722 & 8     & 7      & 5        \\
17      & 15.647 & 9.987  & 7.338  & 9     & 7      & 4        \\
18      & 13.311 & 9.802  & 13.220 & 7     & 9      & 4        \\
19      & 14.091 & 6.207  & 4.188  & 6     & 9      & 5        \\
20      & 14.902 & 9.841  & 6.651  & 8     & 8      & 4        \\
21      & 14.084 & 9.976  & 11.592 & 7     & 8      & 5        \\
22      & 14.281 & 9.878  & 6.822  & 7     & 8      & 5        \\
23      & 15.985 & 9.658  & 14.928 & 10    & 6      & 4        \\
24      & 15.826 & 9.960  & 9.902  & 10    & 6      & 4        \\
25      & 15.287 & 9.975  & 9.691  & 9     & 7      & 4        \\
26      & 13.946 & 9.689  & 11.942 & 9     & 7      & 4        \\
27      & 15.586 & 9.843  & 11.759 & 9     & 7      & 4        \\
28      & 19.997 & 9.743  & 14.043 & 9     & 8      & 3        \\
29      & 15.750 & 12.970 & 13.227 & 9     & 7      & 4        \\
30      & 14.977 & 9.922  & 13.238 & 8     & 8      & 4        \\
31      & 14.717 & 9.983  & 9.602  & 9     & 7      & 4        \\
32      & 16.228 & 6.057  & 9.794  & 9     & 8      & 3        \\
33      & 14.325 & 9.900  & 4.573  & 8     & 7      & 5        \\
34      & 15.218 & 9.932  & 9.953  & 9     & 7      & 4        \\
35      & 14.997 & 9.596  & 11.532 & 7     & 8      & 5        \\
36      & 14.717 & 9.987  & 14.909 & 7     & 10     & 3        \\
37      & 15.076 & 9.985  & 11.289 & 8     & 7      & 5        \\
38      & 14.174 & 9.850  & 7.219  & 7     & 8      & 5        \\
39      & 15.861 & 9.841  & 14.918 & 10    & 6      & 4        \\
40      & 15.995 & 9.784  & 14.837 & 10    & 6      & 4        \\
41      & 14.637 & 6.661  & 6.218  & 7     & 8      & 5        \\
42      & 14.900 & 9.928  & 11.405 & 8     & 7      & 5        \\
43      & 16.652 & 9.960  & 16.463 & 7     & 9      & 4        \\
44      & 15.219 & 9.994  & 9.958  & 9     & 7      & 4        \\
45      & 15.168 & 9.960  & 10.971 & 8     & 7      & 5        \\
46      & 15.347 & 6.324  & 3.722  & 7     & 10     & 3        \\
47      & 19.997 & 9.743  & 14.043 & 9     & 8      & 3        \\
48      & 14.882 & 9.308  & 5.265  & 7     & 10     & 3        \\
49      & 15.189 & 9.981  & 9.894  & 9     & 7      & 4        \\
50      & 14.575 & 9.500  & 4.925  & 8     & 7      & 5        \\
\midrule
Average & 15.205 & 9.558  & 10.158 & 8.18  & 7.56   & 4.26    \\
\bottomrule
\end{tabular}
\end{table}


\begin{table}[ht]
\caption{Results for Player 3 -- Random Estimated Costs (CHR)}
\label{tab:chro_3}
\begin{tabular}{rrrrrrr}
\toprule
\multicolumn{1}{r}{Run} &
\multicolumn{3}{c}{Utilities}    &
\multicolumn{3}{c}{Performance}    \\ 
\cmidrule(lr){2-4}
\cmidrule(lr){5-7}
       & $U_r$    & $U_p$    & $U_s$    & \#win  & \#draw & \#defeat \\
\midrule
1                       & 13.740 & 9.705  & 12.345 & 7     & 8      & 5        \\
2                       & 14.087 & 14.862 & 11.889 & 9     & 6      & 5        \\
3                       & 15.194 & 15.668 & 12.500 & 10    & 6      & 4        \\
4                       & 14.236 & 9.965  & 13.760 & 9     & 8      & 3        \\
5                       & 13.540 & 9.987  & 12.323 & 6     & 10     & 4        \\
6                       & 13.625 & 9.763  & 9.938  & 9     & 7      & 4        \\
7                       & 11.586 & 9.619  & 11.092 & 6     & 9      & 5        \\
8                       & 12.797 & 9.776  & 6.243  & 8     & 8      & 4        \\
9                       & 12.236 & 9.826  & 7.403  & 7     & 8      & 5        \\
10                      & 12.866 & 9.665  & 3.433  & 8     & 7      & 5        \\
11                      & 18.338 & 9.749  & 16.961 & 9     & 8      & 3        \\
12                      & 13.199 & 9.960  & 9.909  & 8     & 8      & 4        \\
13                      & 12.028 & 9.833  & 7.950  & 7     & 8      & 5        \\
14                      & 13.474 & 9.806  & 9.996  & 9     & 7      & 4        \\
15                      & 14.089 & 9.947  & 14.594 & 10    & 6      & 4        \\
16                      & 13.537 & 9.976  & 9.928  & 9     & 7      & 4        \\
17                      & 12.286 & 9.826  & 7.756  & 7     & 8      & 5        \\
18                      & 13.775 & 14.918 & 11.935 & 7     & 9      & 4        \\
19                      & 14.501 & 14.990 & 12.425 & 6     & 9      & 5        \\
20                      & 13.521 & 9.889  & 9.374  & 9     & 7      & 4        \\
21                      & 14.871 & 9.886  & 13.275 & 6     & 9      & 5        \\
22                      & 11.454 & 9.994  & 7.765  & 7     & 8      & 5        \\
23                      & 12.444 & 9.960  & 7.566  & 7     & 8      & 5        \\
24                      & 13.737 & 9.959  & 11.457 & 9     & 7      & 4        \\
25                      & 14.271 & 9.982  & 11.175 & 8     & 7      & 5        \\
26                      & 12.497 & 9.996  & 7.856  & 7     & 8      & 5        \\
27                      & 13.911 & 5.955  & 6.613  & 8     & 9      & 3        \\
28                      & 15.236 & 14.945 & 15.767 & 8     & 8      & 4        \\
29                      & 13.526 & 9.991  & 9.950  & 9     & 7      & 4        \\
30                      & 13.543 & 9.647  & 9.975  & 9     & 7      & 4        \\
31                      & 14.071 & 5.900  & 6.341  & 8     & 9      & 3        \\
32                      & 13.740 & 9.825  & 7.419  & 9     & 7      & 4        \\
33                      & 13.353 & 9.992  & 9.800  & 9     & 7      & 4        \\
34                      & 14.471 & 14.791 & 11.022 & 7     & 9      & 4        \\
35                      & 13.548 & 6.574  & 14.925 & 8     & 8      & 4        \\
36                      & 13.672 & 9.977  & 9.928  & 10    & 6      & 4        \\
37                      & 13.798 & 9.746  & 14.372 & 6     & 10     & 4        \\
38                      & 13.904 & 9.805  & 14.841 & 10    & 6      & 4        \\
39                      & 13.270 & 9.262  & 13.232 & 8     & 8      & 4        \\
40                      & 14.102 & 9.877  & 14.890 & 10    & 6      & 4        \\
41                      & 13.328 & 9.830  & 9.782  & 9     & 7      & 4        \\
42                      & 12.690 & 14.938 & 10.930 & 6     & 11     & 3        \\
43                      & 13.234 & 9.930  & 6.402  & 8     & 8      & 4        \\
44                      & 12.342 & 9.686  & 7.297  & 7     & 8      & 5        \\
45                      & 13.552 & 9.961  & 9.745  & 9     & 7      & 4        \\
46                      & 12.840 & 9.690  & 9.765  & 8     & 8      & 4        \\
47                      & 13.640 & 9.832  & 9.687  & 9     & 7      & 4        \\
48                      & 13.576 & 9.938  & 9.756  & 9     & 7      & 4        \\
49                      & 13.338 & 5.930  & 4.055  & 7     & 10     & 3        \\
50                      & 11.620 & 9.836  & 3.150  & 8     & 8      & 4        \\
\midrule
Average                 & 13.525 & 10.267 & 10.210 & 8.06  & 7.78   & 4.16     \\
Error Square of Average & 2.823  & 0.503  & 0.003  & 0.014 & 0.048  & 0.010  \\
\bottomrule
\end{tabular}
\end{table}

\end{document}